\PassOptionsToPackage{table}{xcolor}
\PassOptionsToPackage{sort&compress}{natbib}
\documentclass{ifm-tr}

\usepackage{amsmath,amssymb}
\usepackage{graphicx}
\usepackage{xspace}
\usepackage{afterpage}                 %
\setcitestyle{numbers,square,comma}   %
\definecolor{reflink}{HTML}{4B2E83}
\hypersetup{linkcolor=reflink,citecolor=reflink,urlcolor=reflink}
\definecolor{defaultgray}{gray}{0.9}
\newcommand{\px}{\,px\xspace}

\makeatletter                          %
\DeclareRobustCommand\onedot{\futurelet\@let@token\@onedot}
\def\@onedot{\ifx\@let@token.\else.\null\fi\xspace}
\def\ie{\emph{i.e}\onedot}
\makeatother

\title{Point Dynamics as Proxy for Robotic World Models}

\author[1]{Dong An}\authorsep{,}
\author[1,2]{Tianyu Liu}\authorsep{,}
\author[1,2]{Haoran Ding}\authorsep{,}
\author[1,2]{Wen Yang}\authorsep{,}
\author[2]{Anqing Duan}\authorsep{\\}
\author[1]{Zhengzhong Liu}\authorsep{,}
\author[1,2]{Eric P. Xing}\authorsep{,}
\author[1]{Guangyi Liu}

\affiliation[1]{Institute of Foundation Models}\affiliationsep{,}
\affiliation[2]{Mohamed bin Zayed University of Artificial Intelligence}

\DeclareRobustCommand{\globeicon}{\raisebox{-0.14em}{\begin{tikzpicture}[x=1em,y=1em,line width=0.065em]
  \draw (0,0) circle (0.42);
  \begin{scope}\clip (0,0) circle (0.42);
    \draw (0,0) ellipse (0.18 and 0.42);
    \draw (-0.5,0) -- (0.5,0);
    \draw (-0.5,0.21) -- (0.5,0.21);
    \draw (-0.5,-0.21) -- (0.5,-0.21);
  \end{scope}
\end{tikzpicture}}}
\metadata[\globeicon\ Project Page]{\url{https://marsaki.github.io/point-dynamics-proxy/}}
\date{November 19, 2026}

\abstract{{\Large\sffamily\bfseries\color{black}Overview}\\[1pt]Robotic world models predict future video frames from the current observation and a sequence of robot actions.
While recent models follow the given actions closely, manipulated objects often collapse or move inconsistently with the action.
We attribute this to the conditioning: an action, or a rendering of the robot, specifies where the robot goes but not how the scene responds, leaving a single video model to infer the interaction dynamics implicitly while also rendering appearance.
We therefore separate dynamics prediction from visual rendering, with 3D point dynamics as the proxy between them.
Specifically, we lift the first frame into 3D points, simulate how they respond to the robot's motion with a learned point dynamics model, and render the simulated points back into pixel-aligned videos that any video model can read.
In controlled experiments on the DROID dataset with two video renderers of different architectures, this point dynamics proxy reduces object motion error by 8.6\% and 6.2\% without degrading image quality, and by 8.1\% on a real bimanual robot.
Moreover, the video follows the simulation: more accurate dynamics yield more accurate object motion, and editing the points steers the video.
This suggests that advances in dynamics modeling can translate into more faithful robotic world models.
}

\begin{document}
\maketitle

\section{Introduction}
\label{sec:intro}

Given the current observation and a sequence of robot actions, a robotic world model predicts the future video frames.
Such models serve as learned surrogates of the real world in robot manipulation, where they are used to evaluate policies, to plan actions and to synthesize training data~\cite{yang2024unisim,zhu2025irasim,li2025worldeval,guo2025ctrlworld}.
These uses impose two requirements on the predicted video.
The first is \emph{action following}: the robot must execute the given actions.
The second is \emph{interaction fidelity}: the objects in contact with the robot must respond as they would in reality, whether they are carried, pushed, dragged, deformed or released, while preserving their shape and identity.

\begin{figure}[t]
\centering
\includegraphics[width=\linewidth]{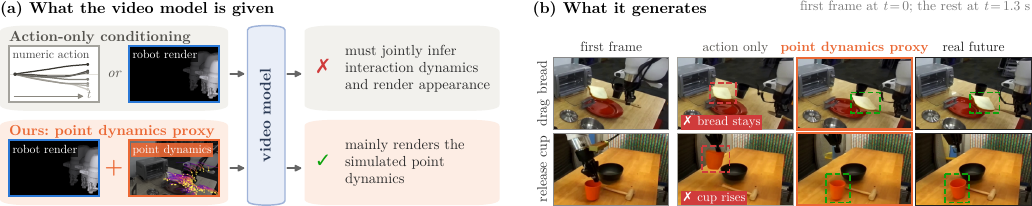}
\caption{\textbf{Action-only conditioning leaves the scene's response to the video model.}
(a)~A simulator predicts how scene points respond, so the video model mainly renders this response.
(b)~Two DROID test clips: objects respond wrongly with action-only conditioning (robot render; red) but not with the proxy (green).}
\label{fig:teaser}
\end{figure}

Early robotic world models encode the action as a numeric vector, such as end-effector poses or joint angles~\cite{yang2024unisim,zhu2025irasim,guo2025ctrlworld}, and leave the video model to infer from pixels both how the robot appears and how the scene responds.
Recent work instead renders the robot into pixel-aligned conditions, such as masks, URDF renders or surface tracks~\cite{alzayer2026mva,kim2026rfwm,li2026hydra0,cao2026tract,azirar2026contactflow}.
This specifies the robot's appearance in every frame and makes action following far more reliable, but interaction fidelity improves much less.
As shown in Fig.~\ref{fig:teaser}b, objects may collapse during contact, as when a slice of bread and the spatula dragging it come apart, or respond inconsistently with the action, as when a released cup rises with the open gripper.
We attribute these failures to the conditioning (Fig.~\ref{fig:teaser}a): an action, numeric or rendered, specifies where the robot goes but not how the scene responds, so a single video model must infer the interaction dynamics implicitly while also rendering appearance.

We therefore separate dynamics prediction from visual rendering.
This separation is inspired by code world models~\cite{chen2026codewm,fang2026coder}, which evolve an explicit world state with a program, render it into a coarse \emph{proxy} video, and let a video model synthesize the final frames; physics-grounded video generation does the same with simulated physics~\cite{liu2024physgen,wang2025physctrl}.
For robot manipulation, the central design choice is the proxy: it must represent arbitrary objects, including deformable and articulated ones, be obtainable from a single real observation, and admit dynamics that can be predicted under robot actions.
Physics simulators require physical parameters, object assets and object-specific solvers that are difficult to obtain in real scenes~\cite{liu2026realwonder,xiao2026vptwin}, and hand-written rules cannot express pushing, dragging or deformation.
We instead use 3D points as the proxy.
Points represent any object without segmentation or a model, can be lifted from a single RGB-D frame, describe the robot and the scene in the same space, and their dynamics can be learned directly from real robot interaction data.

With points as the proxy, the world model first simulates how the points move and then renders them into video (Fig.~\ref{fig:method}).
We lift the first frame into 3D scene points and convert the actions into robot points via forward kinematics.
A learned point dynamics model, the \emph{simulator}, predicts how every scene point responds to the robot's motion; we instantiate it with PointWorld~\cite{huang2026pointworld}.
The simulated points are rendered back, together with the robot, into pixel-aligned proxy videos, which a video model, the \emph{renderer}, reads through its own encoder and turns into realistic frames.
The simulator thus decides what happens and the renderer decides how it looks, and each can be replaced independently.
We study this design in a controlled setting on the DROID dataset~\cite{khazatsky2024droid} with two renderers of different architectures, Wan2.2~\cite{wan2025wan} (DiT) and Ctrl-World~\cite{guo2025ctrlworld} (UNet), comparing four conditions, each giving the video model more explicit information than the last: a numeric action, a robot render, a robot render with depth, and the point dynamics proxy.
We measure interaction fidelity by object motion error, computed from object points tracked in the generated and the real video rather than from image metrics dominated by the background, and repeat the study on a real bimanual robot.

Our experiments show that:
\begin{itemize}
  \item \textbf{Interaction fidelity improves.}
  Against the robot render, the proxy reduces object motion error by 8.6\% and 6.2\% on DROID, without degrading image quality, and by 8.1\% on the real robot.
  The gain concentrates on pushed, dragged and deformed objects (Sec.~\ref{sec:what}), and on the real robot on deformable ones (Sec.~\ref{sec:real}).
  \item \textbf{The video follows the simulation.}
  Object motion error decreases as the simulator improves, from points held static through a hand-written rule to learned simulators, and editing the simulated motion steers the video (Sec.~\ref{sec:follow}).
  \item \textbf{Point dynamics help more as input than as a matched auxiliary loss.}
  With the same data and training budget, point forcing, an auxiliary point-tracking loss following PhysisForcing~\cite{zhang2026physisforcing}, lowers object motion error by 3.4\%, and the simulated point dynamics as input by 8.6\% (Sec.~\ref{sec:implicit}).
\end{itemize}

\section{Related Work}
\label{sec:related}

\afterpage{%
\begin{figure*}[t]
  \centering
  \includegraphics[width=\textwidth]{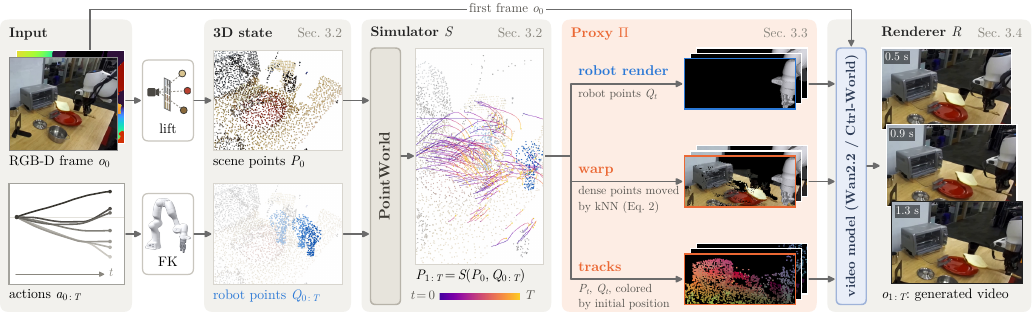}
  \caption{\textbf{Overview.}
  The first frame is lifted into scene points $P_0$, and forward kinematics turns the actions into robot points $Q_{0:T}$ (light to dark blue over time).
  The simulator $S$ predicts how the scene points move (purple to yellow), $\Pi$ renders them with the robot into three pixel-aligned proxy videos, and the renderer $R$ turns the proxy and the first frame into video.
  One of the two cameras is shown.}
  \label{fig:method}
\end{figure*}%
}

\paragraph{Robotic world models for manipulation.}
Robotic world models are increasingly built on large video generation models~\cite{yang2024unisim,zhu2025irasim,guo2025ctrlworld,nvidia2025cosmos,nvidia2025cosmos25,liao2025genieenvisioner,gemini2025veo,jang2025dreamgen,wang2026iws,quevedo2025worldgym,huang2026a2world,nilaksh2026latent}.
What such a model has to infer depends on how the action is given.
A numeric action leaves it to infer both where the robot goes and how the scene responds.
Rendering the robot into the frame as masks, meshes, skeletons, point maps, tracks or flow~\cite{alzayer2026mva,kim2026rfwm,li2026hydra0,cao2026tract,azirar2026contactflow,song2026kineworld,wang2025vap,xu2026kinema4d,kim2025dwm,fu2025robomaster} settles the first question and makes action following far more reliable, but not the second.
A few models also receive a signal about the scene.
Some take it as given, such as simulated depth~\cite{huang2026roborender} or occupancy from reconstruction or simulation~\cite{yang2025orv}; others predict it together with the frames, as next-frame scene flow learned mostly in simulation~\cite{guo2025flowdreamer} or as a low-resolution preview video~\cite{shan2026dvgwm}.
Most work, however, keeps the scene's response inside the video model.
The model is post-trained on counterfactual actions~\cite{li2026dreamtrue}, off-expert actions~\cite{chen2026worldecho} or its own rollouts with reinforcement learning~\cite{bardhan2026persistent}.
Others make it generate depth, point maps or flow along with the frames~\cite{zhen2025tesseract,liu2025geo4dgen,zhao2026rynnworld4d,tong2026pointaction}, or add auxiliary losses on motion and geometry, first for general video generation~\cite{chefer2025videojam,jeong2025track4gen,wu2025geometryforcing,zhang2025videorepa} and recently for robotic world models~\cite{jeon2026vpm,zhang2026physisforcing,wang2026simforcing,chen2026hwm,zhou2026gem4d,li2026s4dlpm}.
We compare one such loss with an explicit input in Sec.~\ref{sec:implicit}.

\paragraph{Simulate, then render.}
Several lines of video generation keep how a world evolves apart from how it looks.
Code world models~\cite{chen2026codewm,fang2026coder} and game generators~\cite{lin2026gamedirector,huang2026progwm} let a program evolve the state and a video model paint it.
Physics-grounded generation does the same with a physics simulator~\cite{liu2024physgen,chen2025physgen3d,xie2024physgaussian,zhang2024physdreamer,li2025wonderplay,liu2026realwonder,foo2026psivg}, a learned physics model~\cite{wang2025physctrl} or an evolving 4D scene state~\cite{wang2026kepler4d}, whose renders, flow or tracks reach the video model as the signals already used to control motion and camera~\cite{geng2025motionprompting,gu2025das,yu2025viewcrafter,ren2025gen3c,gillman2025forceprompting}.
Both stages can also be learned: a language-model backbone evolves a latent world state under actions, and a video diffusion decoder renders it~\cite{panteam2025pan}.
Three concurrent works are closest to ours, as they predict dynamics in 3D before rendering.
DeXplicit models hand--object interaction and needs object labels and canonical shapes~\cite{lee2026dexplicit}.
3D World Simulator fits particle--voxel dynamics and a Gaussian renderer to each multi-camera workspace~\cite{wang2026hpvd}, and VPTwin simulates a digital twin that a VLM builds for each task~\cite{xiao2026vptwin}.
Our intermediate is instead an explicit 3D point state, predicted from the actions by a separately trained point model learned from real robot data, without object models, physical parameters or per-scene capture.

\paragraph{Learned 3D dynamics.}
Learned dynamics models predict how particles, graphs or Gaussians move and deform under robot action, but are mostly fitted to one object or task~\cite{zhang2024gsdyn,zhang2025particlegrid,jiang2025phystwin,huang2025particleformer,lu2025gwm}, and object-centric variants assume segmented rigid or articulated parts~\cite{weng2026ordwm,lee2026dexplicit}.
Scene-level point models drop these assumptions and scale to in-the-wild robot data~\cite{huang2026pointworld,peri2026pw3d,duisterhof2026pointzero,ye2026pointcast,park2026pointwam,bardhan2026depthworld,wang2026unitas,lee2026mu0}, and generative particle models are trained on large simulated corpora~\cite{chen2026physiformer}, including concurrent work~\cite{ren2026gpwm}.
These models predict the evolving 3D state rather than video.
Because the proxy only needs per-point displacements, they could serve as our simulator; we use PointWorld~\cite{huang2026pointworld}, which is public and trained on real robot data, and vary its output at test time in Sec.~\ref{sec:follow}.

\section{Method}
\label{sec:method}

We simulate the scene's response to the robot in 3D and render it into the image space that video models read (Fig.~\ref{fig:method}).

\subsection{Simulate-then-render formulation}
\label{sec:formulation}

A robotic world model predicts future observations $o_{1:T}$ from the first observation $o_0$ and the robot's actions $a_{0:T}$; here $o_0$ holds RGB-D images from calibrated cameras, and $a_t$ is the robot's joint configuration, including the gripper.
We factorize the model into four steps:
\begin{equation}
  Q_{0:T} = \mathrm{FK}\big(a_{0:T}\big), \quad P_{1:T} = S\big(P_0,\, Q_{0:T}\big), \quad
  c_{0:T} = \Pi\big(o_0,\, P_{0:T},\, Q_{0:T}\big), \quad o_{1:T} \sim R\big(o_0,\, c_{0:T}\big).
\label{eq:framework}
\end{equation}
Forward kinematics turns the actions into robot surface points $Q_{0:T}$, the \emph{simulator} $S$ predicts how the scene points $P_0$ lifted from the first frame move in response, $\Pi$ renders the 3D points into pixel-aligned \emph{proxy} videos $c_{0:T}$, and the \emph{renderer} $R$, a video model, turns the proxy into frames.
In this factorization, the actions enter only through kinematics, so only the scene's response has to be predicted.
Existing interfaces use only part of this chain.
Numeric-action models~\cite{yang2024unisim,zhu2025irasim,guo2025ctrlworld} read the actions as a vector, here the end-effector pose given by forward kinematics and the gripper.
Robot-render models~\cite{alzayer2026mva,kim2026rfwm,li2026hydra0} receive a rendering of the robot points $Q_{0:T}$ alone, and a depth condition~\cite{huang2026roborender} adds the scene's geometry, here a depth rendering of $Q_{0:T}$ over the dense scene points held in place, as if $S$ were the identity (Sec.~\ref{sec:setup}).

\subsection{3D state and simulator}
\label{sec:state}

Each camera's depth map is back-projected into the robot base frame, without the pixels covered by the robot at $t{=}0$.
Since the cost of simulation grows with the number of points and the quality of rendering with their density, the simulator receives sparse voxelized points $P_0=\{p_j\}$ merged over cameras, while the proxy uses dense points $\bar P^v=\{\bar p_i\}$, one per pixel of camera $v$ with its color.
Robot points are sampled once on the URDF surface, proportionally to area, and placed at every frame by forward kinematics.
The simulator predicts a displacement $\Delta p_{j,t}$ for every scene point and step, and any model with this interface can serve as $S$ (Sec.~\ref{sec:follow}).
We instantiate $S$ with PointWorld-large~\cite{huang2026pointworld}, a scene-level point dynamics model trained on DROID that reads the gripper points; on the real robot, we fine-tune it on our own recordings (Sec.~\ref{sec:real}).
PointWorld describes each point, voxelized at 1.5\,cm, by DINOv3 features~\cite{simeoni2025dinov3} and predicts its displacement for 10 steps, one every two video frames, which we interpolate linearly in time.
On real data such as DROID, it is trained on \emph{label point flow}, pseudo-labels made of point tracks estimated in the video~\cite{karaev2024cotracker3} and lifted to 3D with depth.

\subsection{Point dynamics proxy}
\label{sec:proxy}

Rendering the points into pixel-aligned videos lets any video model read them through its own VAE.
The proxy sees the simulator only through the simulated points.
It contains the robot render and adds the scene's geometry and its simulated response.
For each camera, $\Pi$ renders three videos by z-buffered point splatting~\cite{zwicker2001splatting,wiles2020synsin}, where each pixel shows the nearest projected point: the robot render $c^{\mathrm{rob}}$, a warp $c^{\mathrm{warp}}$ and a track video $c^{\mathrm{trk}}$.
All cameras are rendered from the same simulated points, so their proxies agree by construction.

\paragraph{Warp.}
Each dense point takes the displacement of its $k$ nearest simulated points $\mathcal{N}_i$ at $t{=}0$ by inverse-distance weighting~\cite{qi2017pointnetpp}:
\begin{equation}
  \Delta \bar p_{i,t} = \frac{\sum_{j\in\mathcal{N}_i} w_{ij}\,\Delta p_{j,t}}{\sum_{j\in\mathcal{N}_i} w_{ij}},
  \qquad
  w_{ij} = \frac{\mathbf{1}\!\left[d_{ij}<r\right]}{d_{ij}+\epsilon},
  \label{eq:knn}
\end{equation}
with $d_{ij}=\lVert\bar p_i-p_j\rVert$, $k{=}8$, $r{=}6$\,cm and $\epsilon{=}1$\,mm; points with no simulated neighbor within $r$ stay static.
The moved dense points and the robot points share one z-buffer, so the arm, the objects it moves and the background occlude each other correctly, and pixels that no point reaches stay black to mark where new content must be generated.

\paragraph{Tracks.}
The track video draws a fixed subset of 4,000 simulated scene points and 1,500 robot points without spatial interpolation, each colored by its initial 3D position in the robot base frame~\cite{gu2025das}.
A dot keeps its color as it moves, so the renderer can follow each point across frames and views, and objects separate from their surroundings by moving differently, without segmentation.
The warp carries appearance but inherits interpolation errors at object boundaries, whereas the tracks show the simulated points directly but sparsely; Sec.~\ref{sec:what} compares the two paths.

\subsection{Renderer}
\label{sec:renderer}

The renderer is a pretrained video model, Wan2.2-TI2V-5B~\cite{wan2025wan} or Ctrl-World~\cite{guo2025ctrlworld} initialized from SVD~\cite{blattmann2023svd}, that receives the first frame of each camera and the three proxy videos, but no text; Ctrl-World, which natively reads numeric actions, keeps this input in every variant.
Each proxy video is encoded by the renderer's frozen VAE, concatenated with the noisy latents along channels and read by a widened first layer whose new weights start at zero, so training starts from the pretrained model.
Each token thus carries the proxy at its own space-time location, and the backbone's attention relates tokens across space, time and views.
The clean first frame supplies appearance, the warp shows where it goes and the tracks how the points move; the renderer fills the black regions and synthesizes the real arm, lighting and contact.
With $\mathcal{E}$ the VAE encoder and $z_\tau$ the noisy latents of the future frames at noise level $\tau$, the denoiser computes
\begin{equation}
  f_\theta\big([\,z_\tau;\ \mathcal{E}(c^{\mathrm{rob}});\ \mathcal{E}(c^{\mathrm{warp}});\ \mathcal{E}(c^{\mathrm{trk}})\,],\ \tau,\ \mathcal{E}(o_0)\big),
  \label{eq:renderer}
\end{equation}
where $[\cdot;\cdot]$ stacks channels, and is trained with the renderer's base denoising objective.
Like its test proxies, its training proxies come from the simulator, errors included, rather than from label point flow.
Architectures and hyperparameters are in Appendix~\ref{sec:suppl-impl}.

\section{Experiments}
\label{sec:experiments}

On DROID and under one controlled protocol, we ask whether point dynamics improve interaction fidelity (Sec.~\ref{sec:what}), whether the generated video follows the simulator (Sec.~\ref{sec:follow}), and whether point dynamics help more as input or as an auxiliary training loss (Sec.~\ref{sec:implicit}).
We then repeat the main comparison on a real robot (Sec.~\ref{sec:real}).

\subsection{Setup}
\label{sec:setup}

\paragraph{Data.}
All models are trained on 326K clips from 41.5K DROID~\cite{khazatsky2024droid} episodes and tested on 1,515 held-out clips from 900 episodes across 13 labs; 300 validation clips are used only to monitor training.
The test clips come from PointWorld's test split, so the simulator is not trained on them either.
A clip is 21 frames at 15\,Hz from the two external cameras, stacked along height at $640{\times}352$ per view, with depth and calibration from PointWorld~\cite{huang2026pointworld}.
No wrist camera and no text are used.

\paragraph{Renderers and conditioning variants.}
We use two renderers and train each with four conditions, each giving it more explicit information than the last: the robot's appearance, then the scene's geometry, then the scene's simulated motion.
Wan2.2-TI2V-5B~\cite{wan2025wan} (\emph{Wan} for short) is a DiT conditioned on the first frame; we train it with a numeric action (8-D end-effector pose and gripper), a robot render, a robot render plus depth (\emph{depth} for short), and the point dynamics proxy.
The depth condition is a second pixel-aligned video that composites the rendered depth of the robot over the depth of the first-frame scene, so it adds the scene's geometry but not its motion (Appendix~\ref{sec:suppl-impl}).
Ctrl-World~\cite{guo2025ctrlworld} is an SVD~\cite{blattmann2023svd} UNet that reads numeric actions through frame-level cross-attention; we retrain it from its SVD initialization in our setting with the numeric action alone and with the same three pixel-aligned conditions added.
The effect of point dynamics is always measured between the robot render and the proxy of the same renderer, which differ only in the proxy's warp and track videos.

\paragraph{Training and inference.}
Every renderer variant uses the same data, the same global sample order, global batch 32 on 8 A100 GPUs, and 30K steps of full fine-tuning from the renderer's pretrained weights (AdamW, learning rate $10^{-5}$, 200 warm-up steps, bfloat16 autocast, no EMA).
Each renderer keeps its own noise schedule, first-frame conditioning and VAE.
Sampling uses 50 steps without guidance.
We report the 30K checkpoint: on the 300 validation clips, every variant with a robot render has nearly converged by 20K steps, and the proxy leads the robot render at every 5K checkpoint (Appendix~\ref{sec:suppl-views}).
On one A100 (batch 1, two views), PointWorld and proxy rendering take 0.93\,s per clip, next to 17.2\,s for generating the video with Wan and 30.9\,s with Ctrl-World; the depth maps are given and not timed.
We do not compare against released checkpoints, because most DROID world models require the wrist view and published numbers use different protocols.

\paragraph{Metrics.}
We measure action following by \emph{robot error} and interaction fidelity by \emph{object motion error}, our main metric.
Both track first-frame query points with CoTracker3~\cite{karaev2024cotracker3} in the real and in the generated video and report the mean pixel distance between the two tracks in the frames where the real track is visible: robot error for up to 150 points inside the robot mask, object motion error for up to 200 points of a 10-pixel grid outside the dilated robot mask that move by at least 5\px in the real video.
The latter is defined on the test clips with at least 10 such points (1,017 and 1,029 of the 1,515 clips in the two views); ``large'' denotes the third with the largest object motion (details in Appendix~\ref{sec:suppl-metrics}).
We also report PSNR, LPIPS and FVD, but they are dominated by the background and the robot, and visual realism is known to be a poor indicator of physical correctness~\cite{motamed2025physicsiq}.
Every metric is computed per view and averaged over the two views, and changes are relative differences of the mean error between two variants, computed before rounding.
All tables report one training seed per variant; the robot render and the proxy are trained with two more seeds (Sec.~\ref{sec:what}).

\subsection{Do point dynamics improve interaction fidelity?}
\label{sec:what}

\begin{table*}[tb]
\centering
\small
\setlength{\tabcolsep}{4pt}
\begin{tabular}{@{}llcccccc@{}}
\toprule
& & \multicolumn{2}{c}{Object motion error$\downarrow$} & & & & Robot \\
\cmidrule(lr){3-4}
Renderer & Condition & all & large & PSNR$\uparrow$ & LPIPS$\downarrow$ & FVD$\downarrow$ & err.$\downarrow$ \\
\midrule
\multirow{4}{*}{\shortstack[l]{Wan2.2\\(DiT)}}
& numeric action & 5.89 & 9.11 & 24.29 & 0.059 & 11.3 & 4.73 \\
& robot render & 5.13 & 7.83 & 26.69 & \textbf{0.045} & 8.0 & 1.82 \\
& robot render + depth & 4.97 & 7.54 & \textbf{26.76} & \textbf{0.045} & 8.1 & 1.79 \\
& point dynamics proxy (ours) & \textbf{4.69} & \textbf{6.96} & 26.75 & \textbf{0.045} & \textbf{7.8} & \textbf{1.77} \\
\midrule
\multirow{4}{*}{\shortstack[l]{Ctrl-World\\(SVD UNet)}}
& numeric action & 5.82 & 9.08 & 24.11 & 0.077 & 22.1 & 4.15 \\
& robot render & 5.30 & 8.12 & 25.69 & 0.066 & 17.1 & 1.76 \\
& robot render + depth & 5.21 & 7.93 & 25.71 & 0.066 & 16.9 & \textbf{1.75} \\
& point dynamics proxy (ours) & \textbf{4.97} & \textbf{7.45} & \textbf{25.77} & \textbf{0.065} & \textbf{16.1} & 1.76 \\
\bottomrule
\end{tabular}
\caption{\textbf{Conditioning variants} (DROID test set, 1,515 clips; every metric averaged over the two camera views).
Object motion and robot errors are in pixels.
The proxy contains the robot render, and Ctrl-World keeps its numeric action in all variants.
Image metrics are bounded by each renderer's VAE and not comparable across renderers; per-view object motion errors are in Appendix~\ref{sec:suppl-views}.}
\label{tab:main}
\end{table*}

\begin{figure*}[tb]
  \centering
  \includegraphics[width=\linewidth]{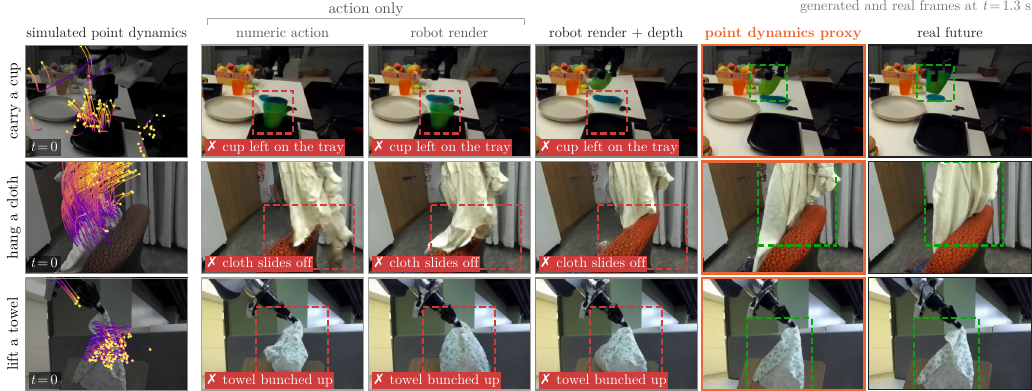}
  \caption{\textbf{The proxy carries the object's response into the video.}
  Test clips (Wan2.2, $t=1.3$\,s, cropped); the left column shows the simulated point trajectories on the first frame.
  Without point dynamics, a carried cup stays on the tray, a cloth slides off the chair and a lifted towel bunches up (red); with the proxy, the objects respond as in the real video (green).}
  \label{fig:qual}
\end{figure*}

\paragraph{Main comparison.}
Simulating point dynamics makes object interaction more faithful on both renderers (Tab.~\ref{tab:main}).
Going from the numeric action to the robot render mostly fixes the robot: robot error drops by 62\% on Wan and 58\% on Ctrl-World, while object motion error drops by only 13.0\% and 9.0\%.
The point dynamics proxy lowers object motion error by another 8.6\% and 6.2\% against the robot render (11.2\% and 8.3\% on large motions), more than half of the robot render's own gain in pixels (0.44 vs.\ 0.76 and 0.33 vs.\ 0.52\px).
Since CoTracker3 also produces the label point flow, we re-measure with AllTracker~\cite{harley2025alltracker}, which uses no pseudo-labels: the proxy then lowers object motion error by 11.5\% and 7.1\%, and SAM~2 object masks~\cite{ravi2025sam2} agree (Appendix~\ref{sec:suppl-metric-val}).
On the whole test set, image metrics barely move (PSNR $+0.06$ and $+0.08$\,dB), and the track error of points that are static in the real video does not rise (0.46\px with and without the proxy on Wan).
In Fig.~\ref{fig:qual}, the models without point dynamics move the arm correctly but get the object wrong; the proxy gets both right.

\begin{table}[tb]
\centering
\begin{minipage}[t]{0.40\linewidth}
\centering
\small
\setlength{\tabcolsep}{6pt}
\begin{tabular}[t]{@{}lcc@{}}
\toprule
Label motion & Wan2.2 & Ctrl-World \\
\midrule
none & $-2.6$ ($-1.4$) & $-2.0$ ($-2.1$) \\
non-rigid & $\mathbf{-12.3}$ ($-8.5$) & $\mathbf{-8.2}$ ($-6.4$) \\
\bottomrule
\end{tabular}
\end{minipage}\hfill
\begin{minipage}[t]{0.56\linewidth}
\centering
\small
\setlength{\tabcolsep}{3pt}
\begin{tabular}[t]{@{}lcccc@{}}
\toprule
& \multicolumn{4}{c}{Object motion error$\downarrow$} \\
\cmidrule(lr){2-5}
Wan2.2 & all & small & large & non-rigid \\
\midrule
robot render (px) & 5.13 & 3.77 & 7.83 & 5.44 \\
+ warp & $-7.0\%$ & $-5.2\%$ & $-8.6\%$ & $-10.7\%$ \\
+ tracks & $-6.5\%$ & $-4.5\%$ & $-8.5\%$ & $-9.5\%$ \\
+ warp + tracks (proxy) & $\mathbf{-8.6\%}$ & $\mathbf{-5.8\%}$ & $\mathbf{-11.2\%}$ & $\mathbf{-12.3\%}$ \\
\bottomrule
\end{tabular}
\end{minipage}\\
\begin{minipage}[t]{0.40\linewidth}
\caption{\textbf{Source of the gain} (DROID test set).
Change of object motion error (\%) against the robot render and, in parentheses, against robot render + depth.}
\label{tab:analysis}
\end{minipage}\hfill
\begin{minipage}[t]{0.56\linewidth}
\caption{\textbf{Proxy paths} (Wan2.2, DROID test set; one training seed per row).
First row: object motion error of the robot render in pixels; other rows: its relative change.
Small and large: the lower two thirds and the largest third of object motion.}
\label{tab:paths}
\end{minipage}
\end{table}

\paragraph{Where the gain comes from.}
Geometry alone does not explain the gain, which lies mostly in non-rigid responses.
The proxy does use the scene's depth, but the depth condition lowers object motion error by only 3.0\% on Wan and 1.7\% on Ctrl-World against the robot render, and the proxy remains 5.7\% and 4.6\% below depth.
We further sort the test clips by their label point flow.
A clip has no labeled motion if fewer than 20 label points move by more than 2\,cm within 15\,cm of the gripper's path, although other points may still move in the video.
A clip with labeled motion has a non-rigid response if fewer than 80\% of these moving label points follow the gripper rigidly, as in pushing, dragging, opening, folding and releasing.
The gain concentrates on the last group (Tab.~\ref{tab:analysis}), where object motion error drops by 12.3\% on Wan and 8.2\% on Ctrl-World, as much as the robot render gains over the numeric action on the same clips (11.9\% and 8.3\%), while clips without labeled motion change little.
This is the division one would expect: the robot render shows where the gripper goes, and with it any object it holds firmly, but not whether an object is held (Fig.~\ref{fig:qual}, top), nor how a cloth falls or a slice of bread slides.

\paragraph{Proxy paths.}
The proxy draws the same point dynamics twice, and either path alone carries most of the gain (Tab.~\ref{tab:paths}): the warp alone reduces object motion error by 7.0\% and the tracks alone by 6.5\%, against 8.6\% for both, so the gain comes from the simulated motion rather than from one way of drawing it.
The two are complementary (Sec.~\ref{sec:proxy}): the warp moves the first frame's appearance densely but blurs motion at object boundaries, the tracks show the simulated points exactly but sparsely, and together they are best in every column, by up to about 3 percentage points.

\begin{figure}[!b]
  \centering
  \begin{minipage}[t]{3.25in}
    \includegraphics[width=\linewidth]{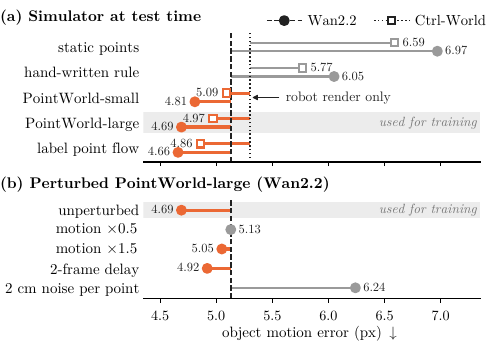}
    \caption{\textbf{The video follows the simulation.}
  Object motion error on the DROID test set when the simulator is swapped (a) or its output perturbed (b) at test time, with the trained renderer fixed.
  Dots left of the vertical lines (robot render only) are better.}
    \label{fig:ladder}
  \end{minipage}\hfill
  \begin{minipage}[t]{3.25in}
    \includegraphics[width=\linewidth]{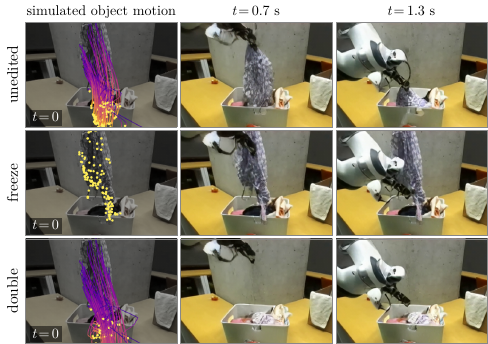}
    \caption{\textbf{Editing the simulated motion steers the video.}
  In this successful example, the simulated motion of the cloth is frozen or doubled at test time, with the robot and the rest of the scene unchanged (Wan2.2).
  Left: the (edited) point trajectories.}
    \label{fig:edit}
  \end{minipage}
\end{figure}

\paragraph{Robustness.}
The gain holds across training seeds and labs.
Pooled over three seeds it is 7.7\% on Wan and 6.7\% on Ctrl-World, with every seed in the same direction, and on each renderer the error decreases in 12 of the 13 DROID labs (Appendix~\ref{sec:suppl-views}).

\subsection{Does the video follow the simulator?}
\label{sec:follow}

If the renderer reads the point dynamics, rather than redrawing what it would have predicted anyway, its video should change whenever the simulation changes.
We test this at inference by swapping, perturbing and editing the simulated motion, with the trained renderer fixed.

\paragraph{Simulator quality.}
Better simulators give better videos on both renderers (Fig.~\ref{fig:ladder}a).
With static points, where every scene point stays in place, object motion error is 49\% higher on Wan and 33\% higher on Ctrl-World than with PointWorld-large, and higher than with the robot render alone, so the renderer follows the motion it is given.
Object motion error then decreases at every step, from a hand-written grasp rule to PointWorld-small, PointWorld-large and label point flow itself.
The rule is a small code simulator: when the gripper closes, the object between the fingers moves rigidly with it until it opens, and everything else stays still.
It closes 40\% (Wan) and 51\% (Ctrl-World) of the gap between static points and PointWorld-large; the rest is pushing, dragging and deformation, which such a rule cannot express.
PointWorld-large nearly matches label point flow (4.69 vs.\ 4.66\px on Wan, 4.97 vs.\ 4.86\px on Ctrl-World).
Label point flow is itself estimated from 2D tracks and stereo depth~\cite{huang2026pointworld}, so it is a reference, not an upper bound.

\paragraph{Perturbations and edits.}
The renderer also follows changes to the simulator's output, whether errors or deliberate edits.
On Wan, coherent errors cost far less than incoherent ones (Fig.~\ref{fig:ladder}b).
Scaling all simulated motion by $0.5{\times}$ or $1.5{\times}$, or delaying it by two frames, costs about 5--10\%, whereas independent 2\,cm noise on each point costs 33\%, because the renderer paints incoherent motion as given.

Edits steer the video (Fig.~\ref{fig:edit}): we edit the predicted motion of one object near the gripper in 200 test clips with non-rigid responses, leaving everything else unchanged, and measure the fraction of tracked object points that end closer to the edited than to the original trajectory, on the 55--76 clips per edit and view in which the edit moves at least five tracked points.
The 200 clips are the non-rigid ones with the largest label motion, and the edited object is the largest connected group of scene points, at 1.5\,cm voxels, that PointWorld-large moves by more than 2\,cm and that start within 15\,cm of the gripper's path.
When the object is frozen, 61\% of its points follow the edit, against 29\% for the robot-render model, which does not see the edit, and the proxy model's error with respect to the edited trajectory drops from 25.2\px without the edit to 12.8\px.
Reversing the horizontal motion or doubling it is followed less often (28\% vs.\ 13\% and 19\% vs.\ 13\%), as the renderer pulls large, unusual motions back toward what it saw in training.
The edits leave the background unchanged and barely affect the robot.

\subsection{Point dynamics as input or as an auxiliary loss?}
\label{sec:implicit}

Point dynamics can also reach a video model through an auxiliary training loss on motion~\cite{chefer2025videojam,jeong2025track4gen,zhang2026physisforcing,wang2026simforcing}, which needs no extra input at inference.
We train another Wan variant on the robot render with \emph{point forcing}: an auxiliary head on block 15 regresses point tracks from intermediate features during training and is discarded at inference, following PhysisForcing~\cite{zhang2026physisforcing} in its loss weight, head learning rate and moving-point targets.
Its targets are the label point flow, the strongest teacher available.
Data, sample order and budget match the other variants.

\begin{table}[!b]
\centering
\begin{minipage}[c]{0.55\linewidth}
\centering
\small
\setlength{\tabcolsep}{3pt}
\begin{tabular}{@{}lccc@{}}
\toprule
& \multicolumn{2}{c}{Object motion error$\downarrow$} & \\
\cmidrule(lr){2-3}
Wan2.2 & all & large & PSNR$\uparrow$ \\
\midrule
robot render & 5.13 & 7.83 & 26.69 \\
robot render + point forcing & 4.95 & 7.54 & \textbf{26.75} \\
point dynamics proxy (ours) & \textbf{4.69} & \textbf{6.96} & \textbf{26.75} \\
\bottomrule
\end{tabular}
\end{minipage}\hfill
\begin{minipage}[c]{0.42\linewidth}
\captionsetup{skip=0pt}
\caption{\textbf{Point dynamics as an auxiliary loss (point forcing) or as input} (test set, averaged over the two views; object motion error in pixels; one training seed per row).}
\label{tab:implicit}
\end{minipage}
\end{table}

\paragraph{Both help; here, input helps more.}
Against the robot render, point forcing lowers object motion error by 3.4\% and raises PSNR by 0.06\,dB (Tab.~\ref{tab:implicit}); with loss weight 1 and the backbone's learning rate for the head, it does not help (Appendix~\ref{sec:suppl-forcing}).
Giving the simulated dynamics as input lowers object motion error by 8.6\%, to 4.69\px against 4.95\px with point forcing.
A likely reason is what the renderer is asked to do: with forcing it must still infer the scene's response from the robot render, whereas the proxy hands it a response computed in 3D.

\subsection{Real robot}
\label{sec:real}

\paragraph{Setup.}
We collect 300 episodes of 15 tasks on a bimanual setup of two WidowX AI arms, recorded at 30\,Hz by a single static RGB-D camera (RealSense D405, $640{\times}480$) that looks down on the table.
The tasks cover pick-and-place, stacking, pushing, folding a shirt or a bag, unfolding a bag, and handing a bag or a pillow from one arm to the other.
We calibrate the camera by fitting the rendered URDF to the robot's silhouettes and depth, and refine the noisy sensor depth with LingBot-Depth~\cite{tan2026lingbotdepth}.
Label point flow follows PointWorld~\cite{huang2026pointworld}, with CoTracker3~\cite{karaev2024cotracker3} tracks lifted to 3D with this depth (Appendix~\ref{sec:suppl-real}).
The renderer is trained on 270 episodes and tested on 431 clips of 30 held-out episodes (two per task); the simulator never sees the test clips.
We fine-tune PointWorld-large on the label point flow for 6K steps, which lowers its error on moving test points from 4.05\,cm zero-shot to 2.24\,cm.
Each Wan variant starts from its DROID model and is fine-tuned for 6K steps with the same recipe on the single view ($640{\times}352$, 21 frames at 15\,Hz).
Object motion error is defined on the 265 test clips with moving objects, 89 of them large.

\begin{figure*}[tb]
  \centering
  \includegraphics[width=\linewidth]{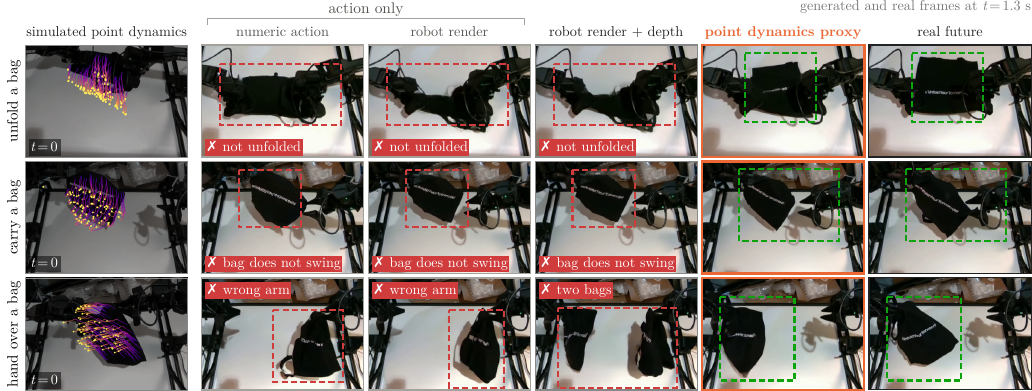}
  \caption{\textbf{The proxy carries the bag's response into the real-robot video.}
  Test clips (Wan2.2, $t=1.3$\,s, cropped); the left column shows the simulated trajectories on the first frame.
  Without point dynamics, the bag is not unfolded, does not swing when carried, or after a handover hangs from the wrong arm or appears twice (red); with the proxy, it responds as in the real video (green).}
  \label{fig:real}
\end{figure*}

\begin{table}[tb]
\centering
\footnotesize
\setlength{\tabcolsep}{4pt}%
\begin{tabular}{@{}lcccccc@{}}
\toprule
& \multicolumn{2}{c}{\shortstack{Object motion\\error$\downarrow$}} & & & & Robot \\
\cmidrule(lr){2-3}
Condition & all & large & PSNR$\uparrow$ & LPIPS$\downarrow$ & FVD$\downarrow$ & err.$\downarrow$ \\
\midrule
numeric action & 3.97 & 5.86 & 27.55 & 0.039 & 23.0 & 2.09 \\
robot render & 2.99 & 4.39 & 30.38 & 0.029 & \textbf{14.5} & 0.74 \\
robot render + depth & 2.87 & 4.13 & \textbf{30.56} & \textbf{0.028} & 14.9 & \textbf{0.72} \\
point dynamics proxy (ours) & \textbf{2.75} & \textbf{3.80} & 30.49 & \textbf{0.028} & 15.2 & 0.73 \\
\bottomrule
\end{tabular}
\caption{\textbf{Real robot} (431 test clips, one camera; Wan2.2 fine-tuned for 6K steps; errors in pixels at $640{\times}352$).}
\label{tab:real}
\end{table}

\paragraph{Results.}
The ordering found on DROID carries over to the real robot (Tab.~\ref{tab:real}, Fig.~\ref{fig:real}).
Against the numeric action, the robot render lowers robot error by 65\% and object motion error by 25\%.
Against the robot render, the proxy lowers object motion error by another 8.1\%, and by 13.4\% on large motions, close to the 8.6\% and 11.2\% on DROID.
Against robot render + depth, it lowers object motion error by 4.3\%, and by 8.0\% on large motions.
The gain comes mostly from deformable objects.
On the 112 clips of the five bag, cloth and pillow tasks, the proxy lowers object motion error by 12.7\% against the robot render and by 7.6\% against depth, close to the 12.3\% and 8.5\% for non-rigid responses on DROID (Tab.~\ref{tab:analysis}); on the 153 clips of the other ten tasks it changes object motion error by at most 2.1\%.
PSNR and LPIPS barely move, while FVD rises from 14.5 to 15.2.

\section{Conclusion}
\label{sec:conclusion}

Robotic world models that render the robot follow the given action, but still have to guess how the objects respond.
We decouple such a model into a simulator that predicts how 3D scene points respond to the robot and a renderer that turns the prediction into video, connected by a pixel-aligned point dynamics proxy.
On DROID and on a real robot, explicit point dynamics make object interaction more faithful, and the generated video follows the simulator and responds to its edits.
This suggests that progress in 3D point dynamics, whether learned or written as code, can carry over to robotic world models without changing how they render.

\paragraph{Limitations.}
The video is only as faithful as the simulation, since the renderer follows plausible simulated motion, errors included.
The simulator, in turn, learns from label point flow, which is itself estimated from tracking and depth.
Errors thus add up from depth and tracking to the labels, the simulator and the video, and each step leaves room for improvement.
The simulator and the renderer are also trained separately, which keeps each replaceable but prevents the renderer from correcting the simulation; training them jointly is a promising direction.

\bibliographystyle{ieeenat_fullname}
\bibliography{main}

\begin{thebibliography}{97}
\providecommand{\natexlab}[1]{#1}
\providecommand{\url}[1]{\texttt{#1}}
\expandafter\ifx\csname urlstyle\endcsname\relax
  \providecommand{\doi}[1]{doi: #1}\else
  \providecommand{\doi}{doi: \begingroup \urlstyle{rm}\Url}\fi

\bibitem[Alzayer et~al.(2026)Alzayer, Huang, Chen, Luey, Zhang, Agrawala,
  et~al.]{alzayer2026mva}
Hadi Alzayer, Wenlong Huang, Haonan Chen, Christopher Luey, Lvmin Zhang,
  Maneesh Agrawala, et~al.
\newblock Masked visual actions for unified world modeling.
\newblock In \emph{Advances in Neural Information Processing Systems
  (NeurIPS)}, 2026.

\bibitem[Azirar et~al.(2026)Azirar, Pallotta, Nogga, Gall, Behnke, and
  Blum]{azirar2026contactflow}
Sami Azirar, Enrico Pallotta, Jan Nogga, J{\"u}rgen Gall, Sven Behnke, and
  Hermann Blum.
\newblock {ContactFlow}: A video action conditioning that transfers across
  embodiments.
\newblock arXiv preprint arXiv:2607.26579, 2026.

\bibitem[Bardhan et~al.(2026{\natexlab{a}})Bardhan, Drozdik, Sivic, and
  Petrik]{bardhan2026persistent}
Jai Bardhan, Patrik Drozdik, Josef Sivic, and Vladimir Petrik.
\newblock Persistent robot world models: Stabilizing multi-step rollouts via
  reinforcement learning.
\newblock In \emph{European Conference on Computer Vision (ECCV)},
  2026{\natexlab{a}}.

\bibitem[Bardhan et~al.(2026{\natexlab{b}})Bardhan, Sivic, and
  Petrik]{bardhan2026depthworld}
Jai Bardhan, Josef Sivic, and Vladimir Petrik.
\newblock {DepthWorld}: {3D} world model for robot manipulation.
\newblock In \emph{Conference on Robot Learning (CoRL)}, 2026{\natexlab{b}}.

\bibitem[Blattmann et~al.(2023)Blattmann, Dockhorn, Kulal, Mendelevitch,
  Kilian, Lorenz, et~al.]{blattmann2023svd}
Andreas Blattmann, Tim Dockhorn, Sumith Kulal, Daniel Mendelevitch, Maciej
  Kilian, Dominik Lorenz, et~al.
\newblock Stable video diffusion: Scaling latent video diffusion models to
  large datasets.
\newblock arXiv preprint arXiv:2311.15127, 2023.

\bibitem[Cao et~al.(2026)Cao, Ji, Zhang, Ge, Fei-Fei, Wu, and
  Huang]{cao2026tract}
Zhi Cao, Howard Ji, Kevin Zhang, Kuangzhi Ge, Li Fei-Fei, Jiajun Wu, and Huang
  Huang.
\newblock {TrAct}: Bridging robot control and visual prediction with visual
  tracks.
\newblock arXiv preprint arXiv:2608.24101, 2026.

\bibitem[Chefer et~al.(2025)Chefer, Singer, Zohar, Kirstain, Polyak, Taigman,
  Wolf, and Sheynin]{chefer2025videojam}
Hila Chefer, Uriel Singer, Amit Zohar, Yuval Kirstain, Adam Polyak, Yaniv
  Taigman, Lior Wolf, and Shelly Sheynin.
\newblock {VideoJAM}: Joint appearance-motion representations for enhanced
  motion generation in video models.
\newblock In \emph{International Conference on Machine Learning (ICML)}, 2025.

\bibitem[Chen et~al.(2025)Chen, Jiang, Liu, Gupta, Li, Zhao, and
  Wang]{chen2025physgen3d}
Boyuan Chen, Hanxiao Jiang, Shaowei Liu, Saurabh Gupta, Yunzhu Li, Hao Zhao,
  and Shenlong Wang.
\newblock {PhysGen3D}: Crafting a miniature interactive world from a single
  image.
\newblock In \emph{IEEE/CVF Conference on Computer Vision and Pattern
  Recognition (CVPR)}, 2025.

\bibitem[Chen et~al.(2026{\natexlab{a}})Chen, Zhan, Liang, Cui, and
  Zhang]{chen2026hwm}
Kevin Chen, Yu-Wei Zhan, Haoran Liang, Sen Cui, and Changshui Zhang.
\newblock Structured latent dynamics for long-horizon embodied world models.
\newblock arXiv preprint arXiv:2610.24534, 2026{\natexlab{a}}.

\bibitem[Chen et~al.(2026{\natexlab{b}})Chen, Liu, Wu, Guo, Wang,
  et~al.]{chen2026worldecho}
Sixiang Chen, Jiaming Liu, Jixian Wu, Yichen Guo, Tinghao Wang, et~al.
\newblock Do robotic world models really follow actions? {D}iagnosing and
  aligning action-conditioned generation for policy learning.
\newblock arXiv preprint arXiv:2608.24885, 2026{\natexlab{b}}.

\bibitem[Chen et~al.(2026{\natexlab{c}})Chen, Lan, and
  Vedaldi]{chen2026physiformer}
Yiming Chen, Yushi Lan, and Andrea Vedaldi.
\newblock {PhysFormer}: Learning to simulate mechanics in world space.
\newblock In \emph{Advances in Neural Information Processing Systems
  (NeurIPS)}, 2026{\natexlab{c}}.

\bibitem[Chen et~al.(2026{\natexlab{d}})Chen, Lin, and Zhang]{chen2026codewm}
Yiwen Chen, Guosheng Lin, and Chi Zhang.
\newblock Code world model: Coding agent as world brain.
\newblock arXiv preprint arXiv:2608.25927, 2026{\natexlab{d}}.

\bibitem[Duisterhof et~al.(2026)Duisterhof, Zhang, Hung, Wen, Birchfield, Li,
  Ramanan, and Ichnowski]{duisterhof2026pointzero}
Bardienus~P. Duisterhof, Kaifeng Zhang, Adam Hung, Bowen Wen, Stan Birchfield,
  Yunzhu Li, Deva Ramanan, and Jeffrey Ichnowski.
\newblock {PointZero}: {3D} point track completion for learning transferable
  {3D} dynamics.
\newblock In \emph{Advances in Neural Information Processing Systems
  (NeurIPS)}, 2026.

\bibitem[Fang et~al.(2026)Fang, Shao, Zhu, Xiao, Chen, Liu,
  et~al.]{fang2026coder}
Zixun Fang, Yawen Shao, Kai Zhu, Jie Xiao, Shihan Chen, Yu Liu, et~al.
\newblock Code plans, diffusion renders: Open-ended generative world modeling.
\newblock arXiv preprint arXiv:2609.26458, 2026.

\bibitem[Foo et~al.(2026)Foo, Huang, Lattas, Moschoglou, Beeler, and
  Theobalt]{foo2026psivg}
Lin~Geng Foo, Mark~He Huang, Alexandros Lattas, Stylianos Moschoglou, Thabo
  Beeler, and Christian Theobalt.
\newblock Physical simulator in-the-loop video generation.
\newblock In \emph{IEEE/CVF Conference on Computer Vision and Pattern
  Recognition (CVPR)}, 2026.

\bibitem[Fu et~al.(2026)Fu, Wang, Liu, Bai, Xu, Wan, Zhang, and
  Lin]{fu2025robomaster}
Xiao Fu, Xintao Wang, Xian Liu, Jianhong Bai, Runsen Xu, Pengfei Wan, Di Zhang,
  and Dahua Lin.
\newblock Learning video generation for robotic manipulation with collaborative
  trajectory control.
\newblock In \emph{International Conference on Learning Representations
  (ICLR)}, 2026.

\bibitem[{Gemini Robotics Team} et~al.(2025){Gemini Robotics Team},
  Choromanski, Devin, Du, Dwibedi, Gao, Jindal, et~al.]{gemini2025veo}
{Gemini Robotics Team}, Krzysztof Choromanski, Coline Devin, Yilun Du,
  Debidatta Dwibedi, Ruiqi Gao, Abhishek Jindal, et~al.
\newblock Evaluating {Gemini} robotics policies in a {Veo} world simulator.
\newblock arXiv preprint arXiv:2512.10675, 2025.

\bibitem[Geng et~al.(2025)Geng, Herrmann, Hur, Cole, Zhang, Pfaff,
  et~al.]{geng2025motionprompting}
Daniel Geng, Charles Herrmann, Junhwa Hur, Forrester Cole, Serena Zhang, Tobias
  Pfaff, et~al.
\newblock Motion prompting: Controlling video generation with motion
  trajectories.
\newblock In \emph{IEEE/CVF Conference on Computer Vision and Pattern
  Recognition (CVPR)}, 2025.

\bibitem[Gillman et~al.(2025)Gillman, Herrmann, Freeman, Aggarwal, Luo, Sun,
  and Sun]{gillman2025forceprompting}
Nate Gillman, Charles Herrmann, Michael Freeman, Daksh Aggarwal, Evan Luo,
  Deqing Sun, and Chen Sun.
\newblock {Force Prompting}: Video generation models can learn and generalize
  physics-based control signals.
\newblock In \emph{Advances in Neural Information Processing Systems
  (NeurIPS)}, 2025.

\bibitem[Gu et~al.(2025)Gu, Yan, Lu, Li, Dou, Si, Dong, Liu, Lin, Liu, Wang,
  and Liu]{gu2025das}
Zekai Gu, Rui Yan, Jiahao Lu, Peng Li, Zhiyang Dou, Chenyang Si, Zhen Dong,
  Qifeng Liu, Cheng Lin, Ziwei Liu, Wenping Wang, and Yuan Liu.
\newblock Diffusion as shader: {3D}-aware video diffusion for versatile video
  generation control.
\newblock In \emph{SIGGRAPH Conference Papers}, 2025.

\bibitem[Guo et~al.(2026{\natexlab{a}})Guo, Ma, Wang, Yang, Liu, and
  Li]{guo2025flowdreamer}
Jun Guo, Xiaojian Ma, Yikai Wang, Min Yang, Huaping Liu, and Qing Li.
\newblock {FlowDreamer}: A {RGB-D} world model with flow-based motion
  representations for robot manipulation.
\newblock \emph{IEEE Robotics and Automation Letters (RA-L)}, 11\penalty0
  (3):\penalty0 2466--2473, 2026{\natexlab{a}}.

\bibitem[Guo et~al.(2026{\natexlab{b}})Guo, Shi, Chen, and
  Finn]{guo2025ctrlworld}
Yanjiang Guo, Lucy~Xiaoyang Shi, Jianyu Chen, and Chelsea Finn.
\newblock {Ctrl-World}: A controllable generative world model for robot
  manipulation.
\newblock In \emph{International Conference on Learning Representations
  (ICLR)}, 2026{\natexlab{b}}.

\bibitem[Harley et~al.(2025)Harley, You, Sun, Zheng, Raghuraman, Gu, Liang,
  Chu, Dave, Tokmakov, You, Ambrus, Fragkiadaki, and
  Guibas]{harley2025alltracker}
Adam~W. Harley, Yang You, Xinglong Sun, Yang Zheng, Nikhil Raghuraman, Yunqi
  Gu, Sheldon Liang, Wen-Hsuan Chu, Achal Dave, Pavel Tokmakov, Suya You, Rares
  Ambrus, Katerina Fragkiadaki, and Leonidas Guibas.
\newblock {AllTracker}: Efficient dense point tracking at high resolution.
\newblock In \emph{IEEE/CVF International Conference on Computer Vision
  (ICCV)}, 2025.

\bibitem[Huang et~al.(2026{\natexlab{a}})Huang, Ai, Chen, Wang, Du, Liu, Yang,
  et~al.]{huang2026roborender}
Huang Huang, Wensi Ai, Ziyu Chen, Youhui Wang, Zijian Du, Yang Liu, Jiaolong
  Yang, et~al.
\newblock {RoboRender}: Robot-oriented video generation for visual sim-to-real
  transfer.
\newblock arXiv preprint arXiv:2610.09254, 2026{\natexlab{a}}.

\bibitem[Huang et~al.(2025)Huang, Chen, Zhang, Sun, and
  Schwager]{huang2025particleformer}
Suning Huang, Qianzhong Chen, Xiaohan Zhang, Jiankai Sun, and Mac Schwager.
\newblock {ParticleFormer}: A {3D} point cloud world model for multi-object,
  multi-material robotic manipulation.
\newblock In \emph{Conference on Robot Learning (CoRL)}, 2025.

\bibitem[Huang et~al.(2026{\natexlab{b}})Huang, Chao, Mousavian, Liu, Fox, Mo,
  and Fei-Fei]{huang2026pointworld}
Wenlong Huang, Yu-Wei Chao, Arsalan Mousavian, Ming-Yu Liu, Dieter Fox, Kaichun
  Mo, and Li Fei-Fei.
\newblock {PointWorld}: Scaling {3D} world models for in-the-wild robotic
  manipulation.
\newblock In \emph{IEEE/CVF Conference on Computer Vision and Pattern
  Recognition (CVPR)}, 2026{\natexlab{b}}.

\bibitem[Huang et~al.(2026{\natexlab{c}})Huang, Zhang, Liu, Zhang, Cheng, and
  Zhang]{huang2026a2world}
Ze Huang, Jiahui Zhang, Hairuo Liu, Chenxi Zhang, Ran Cheng, and Li Zhang.
\newblock Learning transferable dynamics priors from action to world modeling.
\newblock In \emph{European Conference on Computer Vision (ECCV)},
  2026{\natexlab{c}}.

\bibitem[Huang et~al.(2026{\natexlab{d}})Huang, Lin, Lin, Huang, Yu, Niu,
  et~al.]{huang2026progwm}
Zheng-Hui Huang, Guixu Lin, Jiacheng Lin, Yi-Chuan Huang, Ruihan Yu, Muyao Niu,
  et~al.
\newblock Programmable world model.
\newblock arXiv preprint arXiv:2609.10540, 2026{\natexlab{d}}.

\bibitem[Jang et~al.(2025)Jang, Ye, Lin, Xiang, Bjorck, Fang, Hu,
  et~al.]{jang2025dreamgen}
Joel Jang, Seonghyeon Ye, Zongyu Lin, Jiannan Xiang, Johan Bjorck, Yu Fang,
  Fengyuan Hu, et~al.
\newblock {DreamGen}: Unlocking generalization in robot learning through video
  world models.
\newblock In \emph{Conference on Robot Learning (CoRL)}, 2025.

\bibitem[Jeon et~al.(2026)Jeon, Hong, Kim, Lee, Park, Lee, et~al.]{jeon2026vpm}
Byungwoo Jeon, Sunghwan Hong, Dohyeon Kim, Jonghoon Lee, Jungwoo Park, Inhee
  Lee, et~al.
\newblock Learning the {4D} world through point trajectory forecasting.
\newblock arXiv preprint arXiv:2610.28594, 2026.

\bibitem[Jeong et~al.(2025)Jeong, Huang, Ye, Mitra, and
  Ceylan]{jeong2025track4gen}
Hyeonho Jeong, Chun-Hao~Paul Huang, Jong~Chul Ye, Niloy Mitra, and Duygu
  Ceylan.
\newblock {Track4Gen}: Teaching video diffusion models to track points improves
  video generation.
\newblock In \emph{IEEE/CVF Conference on Computer Vision and Pattern
  Recognition (CVPR)}, 2025.

\bibitem[Jiang et~al.(2025)Jiang, Hsu, Zhang, Yu, Wang, and
  Li]{jiang2025phystwin}
Hanxiao Jiang, Hao-Yu Hsu, Kaifeng Zhang, Hsin-Ni Yu, Shenlong Wang, and Yunzhu
  Li.
\newblock {PhysTwin}: Physics-informed reconstruction and simulation of
  deformable objects from videos.
\newblock In \emph{IEEE/CVF International Conference on Computer Vision
  (ICCV)}, 2025.

\bibitem[Karaev et~al.(2025)Karaev, Makarov, Wang, Neverova, Vedaldi, and
  Rupprecht]{karaev2024cotracker3}
Nikita Karaev, Yuri Makarov, Jianyuan Wang, Natalia Neverova, Andrea Vedaldi,
  and Christian Rupprecht.
\newblock {CoTracker3}: Simpler and better point tracking by pseudo-labelling
  real videos.
\newblock In \emph{IEEE/CVF International Conference on Computer Vision
  (ICCV)}, 2025.

\bibitem[Khazatsky et~al.(2024)Khazatsky, Pertsch, Nair, Balakrishna, Dasari,
  Karamcheti, et~al.]{khazatsky2024droid}
Alexander Khazatsky, Karl Pertsch, Suraj Nair, Ashwin Balakrishna, Sudeep
  Dasari, Siddharth Karamcheti, et~al.
\newblock {DROID}: A large-scale in-the-wild robot manipulation dataset.
\newblock In \emph{Robotics: Science and Systems (RSS)}, 2024.

\bibitem[Kim et~al.(2026{\natexlab{a}})Kim, Kim, Cha, and Joo]{kim2026rfwm}
Byungjun Kim, Taeksoo Kim, Hyunsoo Cha, and Hanbyul Joo.
\newblock Robot-factored world models via robot rendering.
\newblock In \emph{Conference on Robot Learning (CoRL)}, 2026{\natexlab{a}}.

\bibitem[Kim et~al.(2026{\natexlab{b}})Kim, Kim, Lee, and Joo]{kim2025dwm}
Byungjun Kim, Taeksoo Kim, Junyoung Lee, and Hanbyul Joo.
\newblock Dexterous world models.
\newblock In \emph{IEEE/CVF Conference on Computer Vision and Pattern
  Recognition (CVPR)}, 2026{\natexlab{b}}.

\bibitem[Lee et~al.(2026{\natexlab{a}})Lee, Xu, Cao, Tian, and
  Liu]{lee2026dexplicit}
Adam Lee, Mutian Xu, Yukang Cao, Shulin Tian, and Ziwei Liu.
\newblock {DeXplicit}: Making physical state explicit in dexterous manipulation
  world models.
\newblock arXiv preprint arXiv:2610.30781, 2026{\natexlab{a}}.

\bibitem[Lee et~al.(2026{\natexlab{b}})Lee, Jung, Lee, Shin, Shahidzadeh, Lee,
  Kim, et~al.]{lee2026mu0}
Seungjae Lee, Yoonkyo Jung, Jusuk Lee, Jonghun Shin, Amir~Hossein Shahidzadeh,
  Yao-Chih Lee, H.~Jin Kim, et~al.
\newblock {$\mu_0$}: A scalable {3D} interaction-trace world model.
\newblock In \emph{Conference on Robot Learning (CoRL)}, 2026{\natexlab{b}}.

\bibitem[Li et~al.(2026{\natexlab{a}})Li, Wen, Zhu, Wang, Du, Li,
  et~al.]{li2026hydra0}
Hongyu Li, Bowen Wen, Xinghao Zhu, Yixuan Wang, Yilun Du, Yunzhu Li, et~al.
\newblock {Hydra-0}: Action flow for generalist world modeling and control.
\newblock arXiv preprint arXiv:2608.18077, 2026{\natexlab{a}}.

\bibitem[Li et~al.(2026{\natexlab{b}})Li, Li, Liu, Liu, Sun, Pan,
  et~al.]{li2026dreamtrue}
Junyan Li, Ruizhi Li, Yu Liu, Xiangshuo Liu, Mingchao Sun, Hongyu Pan, et~al.
\newblock {DreamTrue}: Action-faithful robot world model with counterfactual
  post-training.
\newblock arXiv preprint arXiv:2610.12468, 2026{\natexlab{b}}.

\bibitem[Li et~al.(2025{\natexlab{a}})Li, Zhu, Wen, Shen, and
  Xu]{li2025worldeval}
Yaxuan Li, Yichen Zhu, Junjie Wen, Chaomin Shen, and Yi Xu.
\newblock {WorldEval}: World model as real-world robot policies evaluator.
\newblock arXiv preprint arXiv:2505.19017, 2025{\natexlab{a}}.

\bibitem[Li et~al.(2025{\natexlab{b}})Li, Yu, Liu, Yang, Herrmann, Wetzstein,
  and Wu]{li2025wonderplay}
Zizhang Li, Hong-Xing Yu, Wei Liu, Yin Yang, Charles Herrmann, Gordon
  Wetzstein, and Jiajun Wu.
\newblock {WonderPlay}: Dynamic {3D} scene generation from a single image and
  actions.
\newblock In \emph{IEEE/CVF International Conference on Computer Vision
  (ICCV)}, 2025{\natexlab{b}}.

\bibitem[Li et~al.(2026{\natexlab{c}})Li, Wu, Han, Cai, and Du]{li2026s4dlpm}
Zhiyi Li, Peilin Wu, Xiaoshen Han, Ruojin Cai, and Yilun Du.
\newblock Structured {4D} latent predictive model for robot planning.
\newblock In \emph{International Conference on Machine Learning (ICML)},
  2026{\natexlab{c}}.

\bibitem[Liao et~al.(2026)Liao, Zhou, Huang, Yang, Chen, Jiang, Hu,
  et~al.]{liao2025genieenvisioner}
Yue Liao, Pengfei Zhou, Siyuan Huang, Donglin Yang, Shengcong Chen, Yuxin
  Jiang, Yue Hu, et~al.
\newblock {Genie Envisioner}: A unified world foundation platform for robotic
  manipulation.
\newblock In \emph{International Conference on Learning Representations
  (ICLR)}, 2026.

\bibitem[Lin et~al.(2026)Lin, Deng, Wu, Wen, and Jin]{lin2026gamedirector}
Zijun Lin, Zhiyang Deng, Yuzhe Wu, Bihan Wen, and Yeying Jin.
\newblock {GameDirector}: Decoupling gameplay logic from rendering for
  player-configurable game world models.
\newblock arXiv preprint arXiv:2609.25652, 2026.

\bibitem[Liu et~al.(2024)Liu, Ren, Gupta, and Wang]{liu2024physgen}
Shaowei Liu, Zhongzheng Ren, Saurabh Gupta, and Shenlong Wang.
\newblock {PhysGen}: Rigid-body physics-grounded image-to-video generation.
\newblock In \emph{European Conference on Computer Vision (ECCV)}, 2024.

\bibitem[Liu et~al.(2026{\natexlab{a}})Liu, Chen, Li, Wang, Yu, and
  Wu]{liu2026realwonder}
Wei Liu, Ziyu Chen, Zizhang Li, Yue Wang, Hong-Xing Yu, and Jiajun Wu.
\newblock {RealWonder}: Real-time physical action-conditioned video generation.
\newblock arXiv preprint arXiv:2603.05449, 2026{\natexlab{a}}.

\bibitem[Liu et~al.(2026{\natexlab{b}})Liu, Li, Cousineau, Feng, Burchfiel, and
  Song]{liu2025geo4dgen}
Zeyi Liu, Shuang Li, Eric Cousineau, Siyuan Feng, Benjamin Burchfiel, and
  Shuran Song.
\newblock Geometry-aware {4D} video generation for robot manipulation.
\newblock In \emph{International Conference on Learning Representations
  (ICLR)}, 2026{\natexlab{b}}.

\bibitem[Lu et~al.(2025)Lu, Jia, Li, Chen, Wang, Tang, and Huang]{lu2025gwm}
Guanxing Lu, Baoxiong Jia, Puhao Li, Yixin Chen, Ziwei Wang, Yansong Tang, and
  Siyuan Huang.
\newblock {GWM}: Towards scalable {Gaussian} world models for robotic
  manipulation.
\newblock In \emph{IEEE/CVF International Conference on Computer Vision
  (ICCV)}, 2025.

\bibitem[Motamed et~al.(2026)Motamed, Culp, Swersky, Jaini, and
  Geirhos]{motamed2025physicsiq}
Saman Motamed, Laura Culp, Kevin Swersky, Priyank Jaini, and Robert Geirhos.
\newblock Do generative video models understand physical principles?
\newblock In \emph{IEEE/CVF Winter Conference on Applications of Computer
  Vision (WACV)}, 2026.

\bibitem[Nilaksh et~al.(2026)Nilaksh, Jha, Zholus, and
  Chandar]{nilaksh2026latent}
Nilaksh, Saurav Jha, Artem Zholus, and Sarath Chandar.
\newblock Reconstruction or semantics? {W}hat makes a latent space useful for
  robotic world models.
\newblock In \emph{Advances in Neural Information Processing Systems (NeurIPS),
  Evaluations and Datasets Track}, 2026.

\bibitem[{NVIDIA} et~al.(2025{\natexlab{a}}){NVIDIA}, Agarwal, Ali, Bala,
  Balaji, Barker, Cai, et~al.]{nvidia2025cosmos}
{NVIDIA}, Niket Agarwal, Arslan Ali, Maciej Bala, Yogesh Balaji, Erik Barker,
  Tiffany Cai, et~al.
\newblock Cosmos world foundation model platform for physical {AI}.
\newblock arXiv preprint arXiv:2501.03575, 2025{\natexlab{a}}.

\bibitem[{NVIDIA} et~al.(2025{\natexlab{b}}){NVIDIA}, Ali, Bai, Bala, Balaji,
  Blakeman, Cai, et~al.]{nvidia2025cosmos25}
{NVIDIA}, Arslan Ali, Junjie Bai, Maciej Bala, Yogesh Balaji, Aaron Blakeman,
  Tiffany Cai, et~al.
\newblock World simulation with video foundation models for physical {AI}.
\newblock arXiv preprint arXiv:2511.00062, 2025{\natexlab{b}}.

\bibitem[{PAN Team} et~al.(2025){PAN Team}, Liu, Gu, Deng, Liu,
  et~al.]{panteam2025pan}
{PAN Team}, Zihan Liu, Yi Gu, Mingkai Deng, Guangyi Liu, et~al.
\newblock {PAN}: A world model for general, actionable, and long-horizon world
  simulation.
\newblock arXiv preprint arXiv:2511.09057, 2025.

\bibitem[Park et~al.(2026)Park, Kim, Park, Kwon, Shukla, Sim, Shin, and
  Cho]{park2026pointwam}
Chunghyun Park, Beomjun Kim, Seungcheol Park, Heeseung Kwon, Yashu Shukla,
  Seunghoon Sim, Jinwoo Shin, and Minsu Cho.
\newblock {PointWAM}: {3D} world action modeling for dexterous robotic
  manipulation.
\newblock arXiv preprint arXiv:2610.02840, 2026.

\bibitem[Peri et~al.(2026)Peri, Nguyen, Kim, Fuxin, and Lee]{peri2026pw3d}
Skand Peri, Hung Nguyen, Chanho Kim, Li Fuxin, and Stefan Lee.
\newblock {3D} point world models: Point completion enables more accurate
  dynamics learning.
\newblock arXiv preprint arXiv:2607.00148, 2026.

\bibitem[Qi et~al.(2017)Qi, Yi, Su, and Guibas]{qi2017pointnetpp}
Charles~R. Qi, Li Yi, Hao Su, and Leonidas~J. Guibas.
\newblock {PointNet++}: Deep hierarchical feature learning on point sets in a
  metric space.
\newblock In \emph{Advances in Neural Information Processing Systems
  (NeurIPS)}, 2017.

\bibitem[Quevedo et~al.(2026)Quevedo, Sharma, Sun, Suryavanshi, Liang, and
  Yang]{quevedo2025worldgym}
Julian Quevedo, Ansh~Kumar Sharma, Yixiang Sun, Varad Suryavanshi, Percy Liang,
  and Sherry Yang.
\newblock {WorldGym}: World model as an environment for policy evaluation.
\newblock In \emph{International Conference on Learning Representations
  (ICLR)}, 2026.

\bibitem[Ravi et~al.(2025)Ravi, Gabeur, Hu, Hu, Ryali, Ma, Khedr, R{\"a}dle,
  Rolland, Gustafson, Mintun, Pan, Alwala, Carion, Wu, Girshick, Doll{\'a}r,
  and Feichtenhofer]{ravi2025sam2}
Nikhila Ravi, Valentin Gabeur, Yuan-Ting Hu, Ronghang Hu, Chaitanya Ryali,
  Tengyu Ma, Haitham Khedr, Roman R{\"a}dle, Chloe Rolland, Laura Gustafson,
  Eric Mintun, Junting Pan, Kalyan~Vasudev Alwala, Nicolas Carion, Chao-Yuan
  Wu, Ross Girshick, Piotr Doll{\'a}r, and Christoph Feichtenhofer.
\newblock {SAM} 2: Segment anything in images and videos.
\newblock In \emph{International Conference on Learning Representations
  (ICLR)}, 2025.

\bibitem[Ren et~al.(2026)Ren, Bronars, Zhu, Tang, Zhang, Zakharov, Tokmakov,
  and Fragkiadaki]{ren2026gpwm}
Robert Ren, Matthew Bronars, He Zhu, Yikai Tang, Jianing Zhang, Sergey
  Zakharov, Pavel Tokmakov, and Katerina Fragkiadaki.
\newblock {GPWM}: Modeling the world as {3D} particle motion.
\newblock arXiv preprint arXiv:2611.08338, 2026.

\bibitem[Ren et~al.(2025)Ren, Shen, Huang, Ling, Lu, Nimier-David, M{\"u}ller,
  Keller, Fidler, and Gao]{ren2025gen3c}
Xuanchi Ren, Tianchang Shen, Jiahui Huang, Huan Ling, Yifan Lu, Merlin
  Nimier-David, Thomas M{\"u}ller, Alexander Keller, Sanja Fidler, and Jun Gao.
\newblock {GEN3C}: {3D}-informed world-consistent video generation with precise
  camera control.
\newblock In \emph{IEEE/CVF Conference on Computer Vision and Pattern
  Recognition (CVPR)}, pages 6121--6132, 2025.

\bibitem[Shan et~al.(2026)Shan, Wu, Wang, Zhu, and Wang]{shan2026dvgwm}
Ziyu Shan, Zhenyu Wu, Xiaofeng Wang, Zheng Zhu, and Ziwei Wang.
\newblock {DVG-WM}: Disentangled video generation enables efficient embodied
  world model for robotic manipulation.
\newblock In \emph{European Conference on Computer Vision (ECCV)}, 2026.

\bibitem[Sim{\'e}oni et~al.(2026)Sim{\'e}oni, Vo, Seitzer, Baldassarre, Oquab,
  Jose, et~al.]{simeoni2025dinov3}
Oriane Sim{\'e}oni, Huy~V. Vo, Maximilian Seitzer, Federico Baldassarre, Maxime
  Oquab, Cijo Jose, et~al.
\newblock {DINOv3}.
\newblock \emph{Transactions on Machine Learning Research}, 2026.

\bibitem[Skorokhodov et~al.(2022)Skorokhodov, Tulyakov, and
  Elhoseiny]{skorokhodov2022styleganv}
Ivan Skorokhodov, Sergey Tulyakov, and Mohamed Elhoseiny.
\newblock {StyleGAN-V}: A continuous video generator with the price, image
  quality and perks of {StyleGAN2}.
\newblock In \emph{IEEE/CVF Conference on Computer Vision and Pattern
  Recognition (CVPR)}, 2022.

\bibitem[Song et~al.(2026)Song, Liu, Xu, Liu, Jin, Su,
  et~al.]{song2026kineworld}
Ziying Song, Yuchen Liu, Zhuoran Xu, Ziyang Liu, Jian Jin, Jiangtao Su, et~al.
\newblock {KineWorld}: Action-induced transport fields for embodied world
  modeling.
\newblock arXiv preprint arXiv:2610.06349, 2026.

\bibitem[Tan et~al.(2026)Tan, Sun, Qin, Adai, Fu, Zhou, Zhang, Xu, Zhu, Shen,
  and Xue]{tan2026lingbotdepth}
Bin Tan, Changjiang Sun, Xiage Qin, Hanat Adai, Zelin Fu, Tianxiang Zhou, Han
  Zhang, Yinghao Xu, Xing Zhu, Yujun Shen, and Nan Xue.
\newblock Masked depth modeling for spatial perception.
\newblock In \emph{European Conference on Computer Vision (ECCV)}, 2026.

\bibitem[{Team Wan} et~al.(2025){Team Wan}, Wang, Ai, Wen, Mao, Xie,
  et~al.]{wan2025wan}
{Team Wan}, Ang Wang, Baole Ai, Bin Wen, Chaojie Mao, Chen-Wei Xie, et~al.
\newblock {Wan}: Open and advanced large-scale video generative models.
\newblock arXiv preprint arXiv:2503.20314, 2025.

\bibitem[Tong et~al.(2026)Tong, Jiang, Feng, Liu, and Gu]{tong2026pointaction}
Mutian Tong, Han Jiang, Qiao Feng, Lingjie Liu, and Jiatao Gu.
\newblock {PointAction}: {3D} points as universal action representations for
  robot control.
\newblock arXiv preprint arXiv:2606.03943, 2026.

\bibitem[Wang et~al.(2025{\natexlab{a}})Wang, Chen, Huang, Dou, Liu, Gu, and
  Liu]{wang2025physctrl}
Chen Wang, Chuhao Chen, Yiming Huang, Zhiyang Dou, Yuan Liu, Jiatao Gu, and
  Lingjie Liu.
\newblock {PhysCtrl}: Generative physics for controllable and physics-grounded
  video generation.
\newblock In \emph{Advances in Neural Information Processing Systems
  (NeurIPS)}, 2025{\natexlab{a}}.

\bibitem[Wang et~al.(2026{\natexlab{a}})Wang, Duan, Wu, Liu, and
  Yan]{wang2026kepler4d}
Feiran Wang, Bin Duan, Junyi Wu, Gaowen Liu, and Yan Yan.
\newblock {Kepler4D}: Controllable future video generation via {4D} scene state
  evolution.
\newblock arXiv preprint arXiv:2610.04152, 2026{\natexlab{a}}.

\bibitem[Wang et~al.(2026{\natexlab{b}})Wang, Chou, Zhang, Wang, Huang, Yu,
  Nabi, and Li]{wang2026hpvd}
Jimmy Wang, Yifan Chou, Kaifeng Zhang, Yixuan Wang, Binghao Huang, Tao Yu,
  Saleh Nabi, and Yunzhu Li.
\newblock Hybrid particle-voxel dynamics for multimodal {3D} world simulation.
\newblock arXiv preprint arXiv:2611.04492, 2026{\natexlab{b}}.

\bibitem[Wang et~al.(2025{\natexlab{b}})Wang, Xu, Dong, Deng, Xiang, Lv,
  et~al.]{wang2025moge2}
Ruicheng Wang, Sicheng Xu, Yue Dong, Yu Deng, Jianfeng Xiang, Zelong Lv, et~al.
\newblock {MoGe-2}: Accurate monocular geometry with metric scale and sharp
  details.
\newblock In \emph{Advances in Neural Information Processing Systems
  (NeurIPS)}, 2025{\natexlab{b}}.

\bibitem[Wang et~al.(2026{\natexlab{c}})Wang, Su, Zhou, Yue, Liu, Lu,
  et~al.]{wang2026unitas}
Ruixiang Wang, Yongyi Su, Wenlve Zhou, Bo Yue, Hengyan Liu, Dekun Lu, et~al.
\newblock {UNITAS}: A 3d-native world action model for embodied manipulation.
\newblock arXiv preprint arXiv:2610.12099, 2026{\natexlab{c}}.

\bibitem[Wang et~al.(2026{\natexlab{d}})Wang, Li, Song, Xie, Zhong, Wu, and
  Peng]{wang2026simforcing}
Xiaodong Wang, Tianle Li, Chuanxin Song, Junliang Xie, Zhanmi Zhong, Suiying
  Wu, and Peixi Peng.
\newblock {SimForcing}: Distilling simulation motion priors into real-domain
  robot world models.
\newblock arXiv preprint arXiv:2610.06598, 2026{\natexlab{d}}.

\bibitem[Wang et~al.(2025{\natexlab{c}})Wang, Wen, Guo, Peng, Qin, Bao, Zhou,
  and Hu]{wang2025vap}
Yuang Wang, Chao Wen, Haoyu Guo, Sida Peng, Minghan Qin, Hujun Bao, Xiaowei
  Zhou, and Ruizhen Hu.
\newblock Precise action-to-video generation through visual action prompts.
\newblock In \emph{IEEE/CVF International Conference on Computer Vision
  (ICCV)}, 2025{\natexlab{c}}.

\bibitem[Wang et~al.(2026{\natexlab{e}})Wang, Syed, Wu, Zhang, Onol, Barreiros,
  Nayyeri, et~al.]{wang2026iws}
Yixuan Wang, Rhythm Syed, Fangyu Wu, Mengchao Zhang, Aykut Onol, Jose
  Barreiros, Hooshang Nayyeri, et~al.
\newblock Interactive world simulator for robot policy training and evaluation.
\newblock In \emph{Robotics: Science and Systems (RSS)}, 2026{\natexlab{e}}.

\bibitem[Weng et~al.(2026)Weng, Zhang, Ou, Mu, Zhao, and Li]{weng2026ordwm}
Yueyang Weng, Xiaopeng Zhang, Yuxi Ou, Yongjin Mu, Chengzhi Zhao, and Yanjie
  Li.
\newblock {ORD-WM}: Learning object-centric rigid dynamics for {3D} world
  modeling in robotic manipulation.
\newblock In \emph{Conference on Robot Learning (CoRL)}, 2026.

\bibitem[Wiles et~al.(2020)Wiles, Gkioxari, Szeliski, and
  Johnson]{wiles2020synsin}
Olivia Wiles, Georgia Gkioxari, Richard Szeliski, and Justin Johnson.
\newblock {SynSin}: End-to-end view synthesis from a single image.
\newblock In \emph{IEEE/CVF Conference on Computer Vision and Pattern
  Recognition (CVPR)}, pages 7465--7475, 2020.

\bibitem[Wu et~al.(2026)Wu, Wu, He, Guo, Ye, Duan, and
  Bian]{wu2025geometryforcing}
Haoyu Wu, Diankun Wu, Tianyu He, Junliang Guo, Yang Ye, Yueqi Duan, and Jiang
  Bian.
\newblock Geometry forcing: Marrying video diffusion and {3D} representation
  for consistent world modeling.
\newblock In \emph{International Conference on Learning Representations
  (ICLR)}, 2026.

\bibitem[Wu et~al.(2024)Wu, Jiang, Wang, Liu, Liu, Qiao, Ouyang, He, and
  Zhao]{wu2024ptv3}
Xiaoyang Wu, Li Jiang, Peng-Shuai Wang, Zhijian Liu, Xihui Liu, Yu Qiao, Wanli
  Ouyang, Tong He, and Hengshuang Zhao.
\newblock Point transformer {V3}: Simpler, faster, stronger.
\newblock In \emph{IEEE/CVF Conference on Computer Vision and Pattern
  Recognition (CVPR)}, pages 4840--4851, 2024.

\bibitem[Xiao et~al.(2026)Xiao, Pan, He, Zhou, and Wang]{xiao2026vptwin}
Zhenghao Xiao, Minting Pan, Nantian He, Dongzhan Zhou, and Yunbo Wang.
\newblock {VPTwin}: Real-sim-real video prediction for robotic manipulation
  planning.
\newblock arXiv preprint arXiv:2609.33104, 2026.

\bibitem[Xie et~al.(2024)Xie, Zong, Qiu, Li, Feng, Yang, and
  Jiang]{xie2024physgaussian}
Tianyi Xie, Zeshun Zong, Yuxing Qiu, Xuan Li, Yutao Feng, Yin Yang, and
  Chenfanfu Jiang.
\newblock {PhysGaussian}: Physics-integrated {3D} {Gaussians} for generative
  dynamics.
\newblock In \emph{IEEE/CVF Conference on Computer Vision and Pattern
  Recognition (CVPR)}, 2024.

\bibitem[Xu et~al.(2026)Xu, Zhang, Liu, Chen, Han, and Liu]{xu2026kinema4d}
Mutian Xu, Tianbao Zhang, Tianqi Liu, Zhaoxi Chen, Xiaoguang Han, and Ziwei
  Liu.
\newblock {Kinema4D}: Kinematic {4D} world modeling for spatiotemporal embodied
  simulation.
\newblock arXiv preprint arXiv:2603.16669, 2026.

\bibitem[Yang et~al.(2024)Yang, Du, Ghasemipour, Tompson, Kaelbling,
  Schuurmans, and Abbeel]{yang2024unisim}
Sherry Yang, Yilun Du, Kamyar Ghasemipour, Jonathan Tompson, Leslie Kaelbling,
  Dale Schuurmans, and Pieter Abbeel.
\newblock Learning interactive real-world simulators.
\newblock In \emph{International Conference on Learning Representations
  (ICLR)}, 2024.

\bibitem[Yang et~al.(2026)Yang, Li, Xu, Wang, Ye, Chen, Qin,
  et~al.]{yang2025orv}
Xiuyu Yang, Bohan Li, Shaocong Xu, Nan Wang, Chongjie Ye, Zhaoxi Chen, Minghan
  Qin, et~al.
\newblock {ORV}: {4D} occupancy-centric robot video generation.
\newblock In \emph{IEEE/CVF Conference on Computer Vision and Pattern
  Recognition (CVPR)}, 2026.

\bibitem[Ye et~al.(2026)Ye, Worobel, Xie, Li, Yu, Hong, and
  Choi]{ye2026pointcast}
Hantao Ye, Ross Worobel, Zhuoli Xie, Mingen Li, Houjian Yu, Youngjin Hong, and
  Changhyun Choi.
\newblock {PointCast}: One world model for rigid, articulated, and deformable
  object manipulation.
\newblock arXiv preprint arXiv:2609.28393, 2026.

\bibitem[Yu et~al.(2025)Yu, Xing, Yuan, Hu, Li, Huang, Gao, Wong, Shan, and
  Tian]{yu2025viewcrafter}
Wangbo Yu, Jinbo Xing, Li Yuan, Wenbo Hu, Xiaoyu Li, Zhipeng Huang, Xiangjun
  Gao, Tien-Tsin Wong, Ying Shan, and Yonghong Tian.
\newblock {ViewCrafter}: Taming video diffusion models for high-fidelity novel
  view synthesis.
\newblock \emph{IEEE Transactions on Pattern Analysis and Machine
  Intelligence}, 2025.

\bibitem[Zhang et~al.(2025{\natexlab{a}})Zhang, Li, Hauser, and
  Li]{zhang2025particlegrid}
Kaifeng Zhang, Baoyu Li, Kris Hauser, and Yunzhu Li.
\newblock Particle-grid neural dynamics for learning deformable object models
  from {RGB-D} videos.
\newblock In \emph{Robotics: Science and Systems (RSS)}, 2025{\natexlab{a}}.

\bibitem[Zhang et~al.(2024{\natexlab{a}})Zhang, Zhang, and Li]{zhang2024gsdyn}
Mingtong Zhang, Kaifeng Zhang, and Yunzhu Li.
\newblock Dynamic {3D} {Gaussian} tracking for graph-based neural dynamics
  modeling.
\newblock In \emph{Conference on Robot Learning (CoRL)}, 2024{\natexlab{a}}.

\bibitem[Zhang et~al.(2026)Zhang, Deng, Sun, Ma, Wang, Du,
  et~al.]{zhang2026physisforcing}
Peiwen Zhang, Yufan Deng, Shangkun Sun, Juncheng Ma, Duomin Wang, Jonas Du,
  et~al.
\newblock {PhysisForcing}: Physics reinforced world simulator for robotic
  manipulation.
\newblock arXiv preprint arXiv:2606.28128, 2026.

\bibitem[Zhang et~al.(2024{\natexlab{b}})Zhang, Yu, Wu, Feng, Zheng, Snavely,
  Wu, and Freeman]{zhang2024physdreamer}
Tianyuan Zhang, Hong-Xing Yu, Rundi Wu, Brandon~Y. Feng, Changxi Zheng, Noah
  Snavely, Jiajun Wu, and William~T. Freeman.
\newblock {PhysDreamer}: Physics-based interaction with {3D} objects via video
  generation.
\newblock In \emph{European Conference on Computer Vision (ECCV)},
  2024{\natexlab{b}}.

\bibitem[Zhang et~al.(2025{\natexlab{b}})Zhang, Liao, Zhang, Meng, Wan, Yan,
  and Cheng]{zhang2025videorepa}
Xiangdong Zhang, Jiaqi Liao, Shaofeng Zhang, Fanqing Meng, Xiangpeng Wan,
  Junchi Yan, and Yu Cheng.
\newblock {VideoREPA}: Learning physics for video generation through relational
  alignment with foundation models.
\newblock In \emph{Advances in Neural Information Processing Systems
  (NeurIPS)}, 2025{\natexlab{b}}.

\bibitem[Zhao et~al.(2026)Zhao, Zhao, Huang, Li, Zhao, and
  Li]{zhao2026rynnworld4d}
Haoyu Zhao, Xingyue Zhao, Siteng Huang, Xin Li, Deli Zhao, and Zhongyu Li.
\newblock {RynnWorld-4D}: {4D} embodied world models for robotic manipulation.
\newblock arXiv preprint arXiv:2607.06559, 2026.

\bibitem[Zhen et~al.(2025)Zhen, Sun, Zhang, Li, Zhou, Du, and
  Gan]{zhen2025tesseract}
Haoyu Zhen, Qiao Sun, Hongxin Zhang, Junyan Li, Siyuan Zhou, Yilun Du, and
  Chuang Gan.
\newblock Learning {4D} embodied world models.
\newblock In \emph{IEEE/CVF International Conference on Computer Vision
  (ICCV)}, 2025.

\bibitem[Zhou et~al.(2026)Zhou, Chen, Zhan, Hua, Chen, Chang, Qu,
  et~al.]{zhou2026gem4d}
Kaichen Zhou, Yuzhen Chen, Fangneng Zhan, Hang Hua, Grace Chen, Xinhai Chang,
  Ao Qu, et~al.
\newblock {GEM-4D}: Geometry-enhanced video world models for robot
  manipulation.
\newblock arXiv preprint arXiv:2605.22882, 2026.

\bibitem[Zhu et~al.(2025)Zhu, Wu, Guo, Liu, Cheang, and Kong]{zhu2025irasim}
Fangqi Zhu, Hongtao Wu, Song Guo, Yuxiao Liu, Chilam Cheang, and Tao Kong.
\newblock {IRASim}: A fine-grained world model for robot manipulation.
\newblock In \emph{IEEE/CVF International Conference on Computer Vision
  (ICCV)}, 2025.

\bibitem[Zwicker et~al.(2001)Zwicker, Pfister, van Baar, and
  Gross]{zwicker2001splatting}
Matthias Zwicker, Hanspeter Pfister, Jeroen van Baar, and Markus Gross.
\newblock Surface splatting.
\newblock In \emph{Proceedings of the 28th Annual Conference on Computer
  Graphics and Interactive Techniques (SIGGRAPH)}, pages 371--378, 2001.

\end{thebibliography}

\clearpage
\beginappendix
This appendix gives the details needed to reproduce the experiments and the results that did not fit in the main text.
Sections~\ref{sec:suppl-proxy}--\ref{sec:suppl-metrics} describe the DROID setup, with every setting in Tab.~\ref{tab:suppl-hparams}.
Sec.~\ref{sec:suppl-metric-val} measures object motion error again with a second point tracker and with object masks.
Sec.~\ref{sec:suppl-views} breaks the results down by view, training seed, checkpoint and lab.
Sections~\ref{sec:suppl-follow}--\ref{sec:suppl-real} detail the edit experiment, point forcing and the real robot.

\section{Simulator and proxy}
\label{sec:suppl-proxy}

The simulator is PointWorld-large~\cite{huang2026pointworld}, a Point Transformer V3~\cite{wu2024ptv3} with about 1B parameters, used frozen from its released DROID checkpoint.
PointWorld-small, used in Sec.~\ref{sec:follow}, is the smaller released DROID checkpoint.
The simulator reads the scene points, voxelized at 1.5\,cm and described by frozen DINOv3 features~\cite{simeoni2025dinov3}, together with the gripper points.
In a single forward pass, it predicts displacements for 10 steps, one every two video frames.
It was trained on DROID label point flow, but not on our test clips, which come from the test split of its release.
The renderers never see a test episode either.
Stereo depth and camera calibration also come from PointWorld's DROID release, for training and test alike.

For the proxy, the depth map is upsampled to the video resolution of $640{\times}352$, and every pixel with a depth between 0.1 and 3.5\,m becomes a dense scene point.
Robot pixels at $t{=}0$ are removed with the rendered robot mask, dilated by $7{\times}7$ pixels.
The simulator receives only the gripper points, while the proxy renders the full arm from 1M points sampled on its URDF surface.
Simulated displacements are interpolated linearly in time to the 15\,Hz frame rate.
Track points are colored by their initial position, through a fixed box of the robot base frame mapped to RGB.

\section{Renderers}
\label{sec:suppl-impl}

Wan2.2-TI2V-5B~\cite{wan2025wan} reads the proxy through a widened patch embedding.
Its first latent frame is the clean first frame at timestep zero.
The latents of the two views are stacked along height into one token sequence, so tokens attend across views.
Ctrl-World~\cite{guo2025ctrlworld} reads the proxy through a widened input convolution and otherwise keeps its released design, including the first-frame latent concatenation and the cross-attention to numeric actions.

\paragraph{Numeric action.}
On DROID, the numeric action has 8 dimensions: the end-effector position, the sine and cosine of its roll, its pitch, its yaw and the gripper.
Each dimension is normalized by the 1st and 99th percentiles of the training set.
Ctrl-World reads it in every variant, through its action cross-attention.
Wan reads it only in the numeric-action variant.
There, the actions of the four video frames in each latent frame are concatenated, and a two-layer MLP maps them to an offset of that latent frame's time embedding.
The last layer of this MLP starts at zero.
In every Wan variant, the text cross-attention receives the embedding of an empty prompt.

\paragraph{Depth variant.}
The robot render + depth variant (\emph{depth} for short) adds one video per camera to the robot render, encoded like the proxy videos.
Its first two channels hold the z-buffer depth of the rendered robot in each frame and the depth of the dense first-frame scene points.
The third holds their per-pixel minimum, \ie the robot composited over the scene.
Depth is mapped linearly from $[0.3,\,2.0]$\,m to $[1,\,0]$, so near surfaces are bright and pixels without geometry are black.
The video is rendered from the same depth maps, calibration and points as the proxy, but its scene points stay in place.

\paragraph{Training.}
All variants of a renderer are trained on the same data in the same order, with the settings of Tab.~\ref{tab:suppl-hparams}.
The warp-only and track-only variants of Tab.~\ref{tab:paths} add a single proxy video to the robot render and are otherwise trained like the proxy.

\begin{table*}[t]
\centering
\footnotesize%
\resizebox{\linewidth}{!}{%
\begin{tabular}{@{}lll@{}}
\toprule
& Wan2.2-TI2V-5B & Ctrl-World \\
\midrule
\multicolumn{3}{@{}l}{\textit{Input}} \\
Clip & \multicolumn{2}{l}{21 frames at 15\,Hz, first frame given; two external views of $640{\times}352$ stacked along height} \\
Latent of both views & $48{\times}6{\times}44{\times}40$ (Wan VAE) & $4{\times}21{\times}88{\times}80$ (SVD VAE) \\
Channels per condition video & $+48$, widened patch embedding & $+4$, widened input convolution \\
Numeric action & numeric-action variant only, via time embedding & every variant, via frame-level cross-attention \\
Text & empty-prompt embedding & none \\
Micro-conditioning & -- & fps 7, motion bucket 127, noise augmentation 0 \\
\midrule
\multicolumn{3}{@{}l}{\textit{Training}} \\
Initialization & Wan2.2-TI2V-5B & SVD image-to-video \\
Trained parameters & all (5.0B) & UNet and action encoder (1.53B) \\
Objective & flow matching, shift 3 & EDM, $P_\mathrm{mean}{=}0.7$, $P_\mathrm{std}{=}1.6$ \\
Loss on & latent frames 2--6 & frames 2--21 \\
Optimizer & \multicolumn{2}{l}{AdamW, learning rate $10^{-5}$, $\beta{=}(0.9,\,0.999)$, weight decay 0.01, gradient clipping 1.0} \\
Schedule & \multicolumn{2}{l}{global batch 32; 200 linear warm-up steps, then constant; 30K steps; no EMA} \\
Precision & \multicolumn{2}{l}{fp32 master weights, bfloat16 autocast} \\
Condition dropout & \multicolumn{2}{l}{10\% for the pixel-aligned conditions together, 10\% for numeric actions independently} \\
Compute per variant & about 270 A100 hours & about 570 A100 hours \\
\midrule
\multicolumn{3}{@{}l}{\textit{Sampling}} \\
Sampler & Euler, 50 steps, shift 3 & Euler, 50 steps, Karras noise levels \\
Guidance and seed & \multicolumn{2}{l}{no guidance; the same fixed seed for every clip} \\
\midrule
\multicolumn{3}{@{}l}{\textit{Points and proxy (both renderers)}} \\
Simulator input & \multicolumn{2}{l}{scene points in 1.5\,cm voxels with DINOv3 features, gripper points; 10 steps, one every 2 frames} \\
Dense points & \multicolumn{2}{l}{one per pixel, depth 0.1--3.5\,m; robot pixels at $t{=}0$ removed ($7{\times}7$ dilated mask)} \\
Robot points & \multicolumn{2}{l}{1M on the URDF surface} \\
Warp & \multicolumn{2}{l}{$k{=}8$, $r{=}6$\,cm, $\epsilon{=}1$\,mm; displacements interpolated linearly to 15\,Hz} \\
Tracks & \multicolumn{2}{l}{4,000 scene and 1,500 robot points, colored by initial position in $[0,1.1]{\times}[-0.7,0.7]{\times}[-0.3,0.8]$\,m} \\
Splat size & \multicolumn{2}{l}{$2{\times}2$ pixels for robot and warp points, $4{\times}4$ pixels for track points} \\
\bottomrule
\end{tabular}
}
\caption{Settings of the DROID experiments. The real-robot changes are given in Sec.~\ref{sec:suppl-real}.}
\label{tab:suppl-hparams}
\end{table*}

\section{Metrics}
\label{sec:suppl-metrics}

Object motion error asks whether scene points that move in the real video move the same way in the generated one.
Its query points lie on a 10-pixel grid in the first real frame, outside the robot mask dilated by $15{\times}15$ pixels, and CoTracker3~\cite{karaev2024cotracker3} tracks them through the 21 real frames.
A point counts as moving if it moves by at least 5\px.
It must also be visible in at least half of the frames and lie on the dilated robot mask in fewer than 30\% of them.
Clips with at least 10 moving points enter the metric (1,017 and 1,029 of the 1,515 clips per view).
In each of them, up to 200 moving points are sampled with a fixed seed and tracked in the generated video as well.
The error is their mean distance to the real track, over the frames 1--20 in which the real track is visible.
``Large'' denotes the third of the clips with the largest mean peak displacement of their moving points, about 20\px and above.
Robot error uses up to 150 points sampled in the first-frame robot mask eroded by $5{\times}5$ pixels.
The static-point check of Sec.~\ref{sec:what} uses up to 150 grid points that move by less than 1\px and never touch the dilated robot mask.
FVD uses the I3D features of StyleGAN-V~\cite{skorokhodov2022styleganv} on the 21 frames at $224{\times}224$.

The label-motion groups of Tab.~\ref{tab:analysis} use the label points that start within 15\,cm of the gripper's path and end more than 2\,cm from where they start.
Points whose displacement lies along either camera ray are left out.
A clip has no labeled motion if fewer than 20 such points remain.
This holds for 335 and 338 of the 1,017 and 1,029 clips with object motion error (ext1, ext2).
The gripper's rigid motion is fitted to the points of its base with the Kabsch algorithm.
A moving label point follows the gripper rigidly if, from the step before it first moves by 1\,cm, its label trajectory stays within 2\,cm of this motion.
A clip is rigid if at least 80\% of its moving label points follow the gripper rigidly (29 and 33 clips) and non-rigid otherwise (653 and 658 clips).

\section{Validation of object motion error}
\label{sec:suppl-metric-val}

Object motion error is measured with CoTracker3, which also produced the label point flow that PointWorld is trained on.
We therefore score the test videos of Tab.~\ref{tab:main} again, once with a second point tracker and once by comparing object masks, without retraining any model.
Both checks rank the four conditions in the same order as Tab.~\ref{tab:main} on both renderers.

\paragraph{Second point tracker.}
AllTracker~\cite{harley2025alltracker} differs from CoTracker3 in design and in training data.
CoTracker3 tracks sparse query points with a transformer and is fine-tuned on real videos pseudo-labeled by other trackers~\cite{karaev2024cotracker3}.
AllTracker estimates dense long-range flow from the first frame to every other frame and is trained without pseudo-labels, on synthetic point-tracking and optical-flow data.
We run it at $640{\times}352$ on the real and the generated videos.
A point counts as visible if its visibility times its confidence is at least 0.6, the threshold of AllTracker's own evaluation.
The query points, clips, large-motion clips and robot points are those of Sec.~\ref{sec:suppl-metrics}.

\begin{table}[h]
\centering
\footnotesize
\setlength{\tabcolsep}{3pt}
\begin{tabular}{@{}llccccc@{}}
\toprule
& & \multicolumn{2}{c}{CoTracker3} & \multicolumn{3}{c}{AllTracker} \\
\cmidrule(lr){3-4}\cmidrule(lr){5-7}
Renderer & Condition & all & large & all & large & robot \\
\midrule
\multirow{5}{*}{Wan2.2}
& numeric action & 5.89 & 9.11 & 4.16 & 7.28 & 5.00 \\
& robot render & 5.13 & 7.83 & 3.58 & 6.22 & 1.88 \\
& robot render + point forcing & 4.95 & 7.54 & 3.41 & 5.94 & 1.88 \\
& robot render + depth & 4.97 & 7.54 & 3.43 & 5.92 & 1.86 \\
& point dynamics proxy (ours) & \textbf{4.69} & \textbf{6.96} & \textbf{3.17} & \textbf{5.36} & \textbf{1.83} \\
\midrule
\multirow{4}{*}{Ctrl-World}
& numeric action & 5.82 & 9.08 & 4.22 & 7.48 & 4.51 \\
& robot render & 5.30 & 8.12 & 3.77 & 6.61 & 1.97 \\
& robot render + depth & 5.21 & 7.93 & 3.63 & 6.28 & \textbf{1.94} \\
& point dynamics proxy (ours) & \textbf{4.97} & \textbf{7.45} & \textbf{3.50} & \textbf{6.08} & 1.95 \\
\midrule
\multirow{3}{*}{Reference}
& Wan VAE reconstruction & 0.95 & 1.24 & 0.55 & 0.78 & 0.45 \\
& SVD VAE reconstruction & 1.46 & 1.88 & 0.91 & 1.32 & 0.72 \\
& copy first frame & 12.43 & 23.35 & 9.15 & 19.23 & 21.37 \\
\bottomrule
\end{tabular}
\caption{Object motion error and robot error measured with AllTracker (pixels; DROID test set, the clips and query points of Tab.~\ref{tab:main}; averaged over the two views).
The CoTracker3 columns repeat Tab.~\ref{tab:main} and Tab.~\ref{tab:implicit}, and the reference rows average those of Tab.~\ref{tab:suppl-views} over the two views.}
\label{tab:suppl-alltracker}
\end{table}

AllTracker gives a lower object motion error on every row of Tab.~\ref{tab:suppl-alltracker}, for example 0.55 instead of 0.95\px for the Wan VAE reconstruction.
Against the robot render, the proxy lowers object motion error by 11.5\% on Wan and 7.1\% on Ctrl-World, and by 13.9\% and 8.1\% on large motions.
With CoTracker3, the same reductions are 8.6\% and 6.2\%, and 11.2\% and 8.3\%.
Against depth, the proxy lowers the error by 7.5\% and 3.6\%, compared with 5.7\% and 4.6\% with CoTracker3.
On the clips with a non-rigid response (Tab.~\ref{tab:analysis}), the reduction against the robot render is 15.8\% on Wan and 9.1\% on Ctrl-World.
Over the three training seeds, the reduction against the robot render is 11.5\%, 9.2\% and 7.8\% on Wan and 7.1\%, 8.2\% and 11.1\% on Ctrl-World, and 9.5\% and 8.8\% pooled.

The other comparisons on Wan keep their direction as well.
Against the robot render, the warp alone lowers object motion error by 9.2\% and the tracks alone by 8.6\% (Tab.~\ref{tab:paths}).
Point forcing lowers it by 4.8\% against the robot render, and the proxy is 7.0\% below point forcing (Tab.~\ref{tab:implicit}).
When the simulator is swapped (Fig.~\ref{fig:ladder}a), the error decreases from static points (5.20\px) through the hand-written rule, PointWorld-small and PointWorld-large to label point flow (3.07\px).
Static points remain worse than the robot render alone (3.58\px).
Among the perturbations (Fig.~\ref{fig:ladder}b), independent noise costs 49\%, and scaling and delay cost 7--12\%.
In the edit experiment (Sec.~\ref{sec:suppl-follow}), 73\% of the points of a frozen object follow the edit, compared with 44\% for the robot-render model.
The error to the edited trajectory drops from 22.7\px without the edit to 12.0\px.
Reversing is followed by 40\% of the points and doubling by 12\%, compared with 25\% and 6\% for the robot-render model, on 64--77 clips per edit, view and video.

To keep CoTracker3 out of the metric entirely, AllTracker can also select the query points, with the rule of Sec.~\ref{sec:suppl-metrics} applied to its own tracks of the real video.
Then 858 and 869 clips have at least 10 moving points (ext1, ext2).
On them, the proxy lowers object motion error against the robot render by 11.0\% on Wan and 8.0\% on Ctrl-World.

\paragraph{Object masks.}
The second check compares object masks, so that no point tracker enters the comparison.
In the first real frame, the query points of object motion error that lie within 20 pixels of each other form groups.
Each group of at least five points prompts SAM~2.1-large~\cite{ravi2025sam2} with up to eight of its points, spread over the group.
The masks are propagated through the real video and, separately, through each generated video from the same prompts.
We keep the objects whose real mask covers at most 5\% of the first frame and whose centroid moves by at least 5\px, which removes walls and table tops.
Then 655 and 683 clips keep at least one object (ext1, ext2).
In each of frames 1--20 with a non-empty real mask, we compute the IoU of the two masks and, where the generated mask is non-empty, the distance between their centroids.
Both are averaged over frames, objects, clips and the two views.

\begin{table}[h]
\centering
\footnotesize
\setlength{\tabcolsep}{3pt}
\begin{tabular}{@{}llccc@{}}
\toprule
Renderer & Condition & IoU$\uparrow$ & IoU, frame 20$\uparrow$ & centroid$\downarrow$ \\
\midrule
\multirow{5}{*}{Wan2.2}
& numeric action & 0.610 & 0.481 & 10.54 \\
& robot render & 0.670 & 0.555 & 9.40 \\
& robot render + point forcing & 0.673 & 0.560 & 9.05 \\
& robot render + depth & 0.678 & 0.568 & 8.99 \\
& point dynamics proxy (ours) & \textbf{0.685} & \textbf{0.575} & \textbf{8.56} \\
\midrule
\multirow{4}{*}{Ctrl-World}
& numeric action & 0.593 & 0.462 & 12.35 \\
& robot render & 0.634 & 0.524 & 11.67 \\
& robot render + depth & 0.639 & 0.528 & 11.46 \\
& point dynamics proxy (ours) & \textbf{0.643} & \textbf{0.536} & \textbf{11.07} \\
\midrule
\multirow{3}{*}{Reference}
& Wan VAE reconstruction & 0.884 & 0.878 & 5.23 \\
& SVD VAE reconstruction & 0.840 & 0.829 & 7.28 \\
& copy first frame & 0.456 & 0.296 & 19.19 \\
\bottomrule
\end{tabular}
\caption{Agreement of SAM~2 object masks between the real and the generated video (DROID test set, 655 and 683 clips; averaged over the two views).
IoU and centroid error (pixels) are averaged over frames 1--20; the middle column uses frame 20 only.}
\label{tab:suppl-sam}
\end{table}

With SAM~2 masks, the four conditions keep their order in all three columns of Tab.~\ref{tab:suppl-sam}.
Against the robot render, the proxy raises the mean IoU from 0.670 to 0.685 on Wan and from 0.634 to 0.643 on Ctrl-World.
It lowers the centroid error against the robot render by 8.9\% and 5.1\%.
Against depth, it lowers the centroid error by 4.7\% and 3.4\%, and on Wan it also improves on point forcing in all three columns.

\section{Per-view results, training seeds, convergence and labs}
\label{sec:suppl-views}

\begin{table}[!ht]
\centering
\begin{minipage}[b]{\dimexpr\linewidth-3.25in\relax}
\centering
\footnotesize
\setlength{\tabcolsep}{2pt}
\begin{tabular}[b]{@{}llcccc@{}}
\toprule
& & \multicolumn{2}{c}{ext1} & \multicolumn{2}{c}{ext2} \\
\cmidrule(lr){3-4}\cmidrule(lr){5-6}
Renderer & Condition & all & large & all & large \\
\midrule
\multirow{4}{*}{Wan2.2}
& numeric action & 5.92 & 9.49 & 5.86 & 8.74 \\
& robot render & 5.21 & 8.27 & 5.05 & 7.40 \\
& robot render + depth & 5.10 & 8.09 & 4.84 & 7.00 \\
& point dynamics proxy (ours) & \textbf{4.82} & \textbf{7.45} & \textbf{4.56} & \textbf{6.46} \\
\midrule
\multirow{4}{*}{Ctrl-World}
& numeric action & 5.83 & 9.34 & 5.81 & 8.82 \\
& robot render & 5.38 & 8.52 & 5.22 & 7.72 \\
& robot render + depth & 5.26 & 8.27 & 5.15 & 7.59 \\
& point dynamics proxy (ours) & \textbf{5.12} & \textbf{8.00} & \textbf{4.82} & \textbf{6.90} \\
\midrule
\multirow{3}{*}{Reference}
& Wan VAE reconstruction & 0.95 & 1.26 & 0.94 & 1.23 \\
& SVD VAE reconstruction & 1.48 & 1.89 & 1.44 & 1.86 \\
& copy first frame & 12.39 & 23.67 & 12.47 & 23.03 \\
\bottomrule
\end{tabular}
\end{minipage}\hfill
\begin{minipage}[b]{3.0in}
\centering
\includegraphics[width=3.0in]{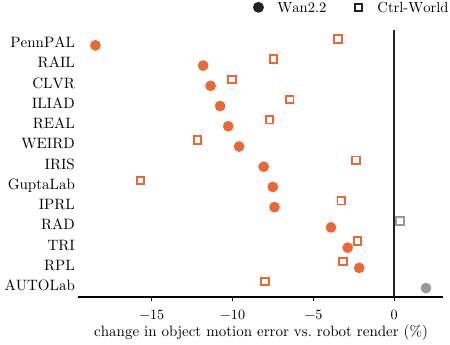}
\end{minipage}\\
\begin{minipage}[t]{\dimexpr\linewidth-3.25in\relax}
\caption{Object motion error per view (pixels; DROID test set; ext1 and ext2 are the two external cameras of DROID).
The reference rows give a lower bound (reconstruction by each renderer's VAE) and a no-motion reference (the first frame repeated).}
\label{tab:suppl-views}
\end{minipage}\hfill
\begin{minipage}[t]{3.0in}
\captionof{figure}{Relative change of object motion error, proxy against robot render, per DROID lab (test set, averaged over the two views).}
\label{fig:suppl-labs}
\end{minipage}
\end{table}

\paragraph{Views.}
Per view (Tab.~\ref{tab:suppl-views}), the proxy reduces object motion error against the robot render by 7.5\% and 9.7\% on Wan and by 4.8\% and 7.6\% on Ctrl-World (ext1, ext2).
VAE reconstruction reaches 38.1 (Wan) and 32.0\,dB (SVD) PSNR, which bounds the image metrics of each renderer.

\paragraph{Training seeds.}
For the comparisons with the robot render, depth and point forcing, we trained two further seeds of each variant.
Against the robot render, the proxy reduces object motion error by 8.6\%, 8.0\% and 6.4\% on Wan and by 6.2\%, 6.7\% and 7.3\% on Ctrl-World for the three seeds.
Pooled over seeds, which averages each clip's error over the three seeds before the comparison, the reduction is 7.7\% and 6.7\%.
Against depth, the proxy reduces the error by 5.7\%, 5.3\% and 7.1\% on Wan and by 4.6\%, 6.1\% and 6.3\% on Ctrl-World, and by 6.1\% and 5.7\% pooled.
On Wan, point forcing reduces the error against the robot render by 3.4\%, 2.0\% and 1.4\%, and by 2.3\% pooled.
The proxy is 5.4\%, 6.1\% and 5.1\% below point forcing, and 5.5\% below it pooled.

\paragraph{Convergence.}
Tab.~\ref{tab:suppl-converge} follows the object motion error of every variant on the 300 validation clips every 5K steps.
The proxy has the lowest error at every checkpoint.
It also stays lowest when each variant is taken at its best checkpoint, with 4.65 vs.\ 4.88\px for the robot render on Wan and 5.15 vs.\ 5.47\px on Ctrl-World.

\paragraph{Labs.}
Against the robot render, the proxy lowers object motion error in 12 of the 13 DROID labs on each renderer (Fig.~\ref{fig:suppl-labs}).

\begin{table}[!hb]
\centering
\begin{minipage}[c]{0.64\linewidth}
\centering
\footnotesize
\setlength{\tabcolsep}{3pt}
\begin{tabular}{@{}llcccccc@{}}
\toprule
Renderer & Condition & 5K & 10K & 15K & 20K & 25K & 30K \\
\midrule
\multirow{4}{*}{Wan2.2}
& numeric action & 13.31 & 8.62 & 7.40 & 6.83 & 6.22 & 5.95 \\
& robot render & 6.23 & 5.68 & 5.23 & 5.10 & 4.88 & 5.35 \\
& robot render + depth & 6.16 & 5.63 & 5.29 & 5.02 & 4.82 & 5.02 \\
& point dynamics proxy (ours) & \textbf{5.45} & \textbf{5.11} & \textbf{4.75} & \textbf{4.92} & \textbf{4.65} & \textbf{4.70} \\
\midrule
\multirow{4}{*}{Ctrl-World}
& numeric action & 11.86 & 7.98 & 7.15 & 6.94 & 6.31 & 6.00 \\
& robot render & 6.47 & 5.94 & 6.02 & 5.54 & 5.47 & 5.51 \\
& robot render + depth & 6.38 & 5.91 & 5.84 & 5.46 & 5.57 & 5.44 \\
& point dynamics proxy (ours) & \textbf{6.23} & \textbf{5.64} & \textbf{5.45} & \textbf{5.31} & \textbf{5.17} & \textbf{5.15} \\
\bottomrule
\end{tabular}
\end{minipage}\hfill
\begin{minipage}[c]{0.33\linewidth}
\captionsetup{skip=0pt}
\caption{Object motion error during training (pixels; 300 validation clips, averaged over the two views; one seed per row).
Columns are training steps.}
\label{tab:suppl-converge}
\end{minipage}
\end{table}

\section{Editing the simulated motion}
\label{sec:suppl-follow}

The edit experiment of Sec.~\ref{sec:follow} uses the 200 non-rigid test clips whose moving label points have the largest 95th percentile of final displacement.
In each clip, we take the PointWorld-large points that move by more than 2\,cm and start within 15\,cm of the gripper's path.
The edited object is their largest 26-connected group at 1.5\,cm voxels.
A group needs at least 20 points; otherwise the next clip in the ranking is taken.
The edits change the object's predicted displacements in the robot base frame.
Freezing keeps the object points at their initial positions, reversing negates their horizontal displacement and keeps the vertical one, and doubling multiplies their displacement by two.
The edited points are rendered into the proxy exactly like the prediction, and the robot render and the rest of the scene are left unchanged.

For scoring, we take the query points of object motion error that lie within 6 pixels of a visible object point in the first frame.
CoTracker3 tracks them in each video.
Each query point has two targets: its position moved by the original and by the edited simulated motion, interpolated as in the warp and projected into the view.
A point counts if it is visible in at least 5 of frames 1--20 and its two targets are at least 2\px apart on average over these frames.
It follows the edit if its track is closer on average to the edited target than to the original one.
A clip and view enters the follow rate if at least 5 of its points count, which leaves 55--76 clips per edit, view and video.
The follow rate is the fraction of counted points that follow the edit, averaged over these clips.
The follow error is the mean distance of their tracks to the edited target.
Against the unedited proxy video, the edits change background error by less than 1\% and PSNR by at most 0.23\,dB.
They change robot error by at most 0.25\px (reverse, ext1, from 2.12\px) and by at most 0.15\px otherwise.

\section{Point forcing}
\label{sec:suppl-forcing}

The forcing head reads block 15 of Wan and regresses the label point flow of moving points.
The variant in Tab.~\ref{tab:implicit} follows PhysisForcing~\cite{zhang2026physisforcing} in its loss weight (0.1), the learning rate of the head ($10^{-4}$, with the same warm-up as the backbone) and its use of moving points only.
Unlike PhysisForcing, it learns from label point flow rather than CoTracker3 tracks and uses a Huber rather than an RMSE loss.
Its batch size and training length follow our protocol (32 and 30K steps, against 128 and 20K).
Our first forcing variant used loss weight 1, the backbone's learning rate $10^{-5}$ for the head and static points weighted by 0.25.
It reached 5.21 / 7.95\px object motion error (all / large) and 26.53\,dB PSNR, on par with no point dynamics ($+1.6\%$).
The recipe of Tab.~\ref{tab:implicit} lowers object motion error by 4.9\% against this first variant.
Sec.~\ref{sec:suppl-views} reports point forcing for three training seeds.

\section{Real-robot details}
\label{sec:suppl-real}

\paragraph{Data.}
Five of the 15 tasks put objects in place: all food on a plate, a cake on a yellow plate, an eggplant into a white box, a lemon into a blue bowl and a red cuboid on a green plate.
Two push a green bowl to a purple bowl and a green cube to a blue cube.
Three stack a purple bowl in a blue bowl, a red cube on a green cube and three bowls.
The other five fold a black bag and a red shirt in half, unfold a black bag, and hand a black bag or a U-shaped pillow from one arm to the other.
Each task has 20 episodes, 176K frames in total.
The robot state is the 14-D joint state of the two arms (six joints and a gripper each).

\paragraph{Calibration.}
No camera calibration was recorded, so we estimate it.
RGB and depth share the optical center, since a free translation between them converges to 4\,mm.
The principal point of the depth image, however, is about 11 pixels lower, so we reproject depth onto the RGB grid.
In 360 frames, we render the arm from the URDF at the logged joint state.
The focal length, principal point and camera pose are fitted jointly to the arm's silhouettes and to the depth on its pixels.
The fitted focal length is about 380 pixels and varies by 0.7\% across three disjoint subsets of frames.
On held-out frames, the depth error on robot pixels is 17--18\,mm and the silhouette IoU is 0.76--0.77.

\paragraph{Depth.}
At about 1\,m, the raw sensor depth on the dark tablecloth fluctuates by centimeters.
We compared raw and filtered sensor depth with MoGe-2~\cite{wang2025moge2} aligned to the sensor and with LingBot-Depth~\cite{tan2026lingbotdepth}, which takes RGB and raw depth, on 232 held-out windows of training episodes.
Three measures score them: the median depth error on robot pixels against the rendered robot, the frame-to-frame jitter of static points, and the change of pairwise distances between points on carried rigid objects.
LingBot-Depth gives 17.9, 12.1 and 4.0\,mm against 17.8, 27.7 and 8.1\,mm for raw depth, while MoGe-2 places the arm 85\,mm off.
We therefore use LingBot-Depth for the labels, the simulator's input points and the depth condition.

\paragraph{Label point flow.}
We follow PointWorld's labeling~\cite{huang2026pointworld}.
CoTracker3~\cite{karaev2024cotracker3} is queried on a dense 3-pixel grid in the first frame, excluding robot pixels.
Each track is lifted to 3D with the median depth of its $5{\times}5$ neighborhood on the same surface.
In every frame, points hidden by the arm, found with the rendered robot mask, are marked invisible and not supervised in that frame.
Outliers are removed with PointWorld's DBSCAN filter.
Two changes account for the single overhead camera.
First, the depth jitter of static points (12\,mm) exceeds the 5\,mm per-step scale of PointWorld's motion weighting.
Static points are therefore pinned to their first-frame position and supervised only in frames where they are visible.
Second, a held object is often hidden by the gripper.
Points whose observed motion follows the gripper are attached to it.
The other points of the same object move with them, by the object's rigid motion fitted per frame with the Kabsch algorithm.
In frames where these points are occluded, they are filled in with that rigid motion.
Points with contradictory evidence, such as a depth change without 2D motion, are left unsupervised after the first frame.

\paragraph{Simulator.}
We start from the released PointWorld-large checkpoint used on DROID and feed it the points of the single camera.
The state normalization is that of the DROID release, with the gripper statistics applied to both grippers.
It is fine-tuned on 8 GPUs with global batch 32 and learning rate $10^{-5}$ for 6K steps, and all other settings follow the release.
Fine-tuning lowers the error on moving points from 4.05 to 2.24\,cm (Tab.~\ref{tab:suppl-real-sim}).

\begin{table}[h]
\centering
\begin{minipage}[c]{0.42\linewidth}
\centering
\small
\setlength{\tabcolsep}{4pt}
\begin{tabular}{@{}lcc@{}}
\toprule
PointWorld & moving$\downarrow$ & static$\downarrow$ \\
\midrule
no motion & 5.07 & -- \\
small, zero-shot & 4.51 & 0.90 \\
large, zero-shot & 4.05 & 1.06 \\
small, fine-tuned & 2.27 & 0.03 \\
large, fine-tuned (used) & \textbf{2.24} & 0.06 \\
\bottomrule
\end{tabular}
\end{minipage}\hfill
\begin{minipage}[c]{0.54\linewidth}
\captionsetup{skip=0pt}
\caption{Simulator error on the 431 real-robot test clips (mean distance to the label point flow over the predicted steps, in cm); moving points are those whose label moves by more than 1\,cm.}
\label{tab:suppl-real-sim}
\end{minipage}
\end{table}

\paragraph{Renderer.}
Each Wan variant is initialized from its 30K DROID model and trained on one view of $640{\times}352$, the lower 352 rows of the $640{\times}480$ frame.
Training follows the DROID recipe of Tab.~\ref{tab:suppl-hparams} on 8 GPUs for 6K steps, and sampling uses 50 steps without guidance.
The numeric action is the position, roll, pitch and yaw of each end effector, obtained by forward kinematics from the measured joints, and the gripper of each arm.
It is normalized by the 1st and 99th percentiles of the training set, and its encoder is trained from scratch because the dimension differs from DROID.
The depth condition is built as on DROID, from the first-frame LingBot-Depth.
The proxy comes from the fine-tuned PointWorld-large.
As on DROID, the training proxies are its predictions on clips it was trained on.

\end{document}